\documentclass{article}

\usepackage{microtype}
\usepackage{graphicx}
\usepackage{subfigure}
\usepackage{booktabs} % for professional tables
\usepackage{multirow}
\usepackage{hyperref}
\usepackage{comment}

\usepackage[accepted]{icml2024}

\usepackage{amsmath}
\usepackage{amssymb}
\usepackage{mathtools}
\usepackage{amsthm}

\usepackage{multicol}

\usepackage[capitalize,noabbrev]{cleveref}

\theoremstyle{plain}

\theoremstyle{definition}

\theoremstyle{remark}

\usepackage[textsize=tiny]{todonotes}

\def \E {\mathrm{E}}

\def \R {\mathbb{R}}

\def \N {\mathcal{N}}

\renewcommand{\algorithmicrequire}{ \textbf{Input:}} %Use Input in the format of Algorithm
\renewcommand{\algorithmicensure}{ \textbf{Output:}} %UseOutput in the format of Algorithm

\icmltitlerunning{Attention as Robust Representation for Time Series Forecasting}

\begin{document}

\twocolumn[
\icmltitle{Attention as Robust Representation for Time Series Forecasting}



\icmlsetsymbol{equal}{*}

\begin{icmlauthorlist}
\icmlauthor{PeiSong Niu}{equal,yyy}
\icmlauthor{Tian Zhou}{equal,yyy}
\icmlauthor{Xue Wang}{yyy}
\icmlauthor{Liang Sun}{yyy}
\icmlauthor{Rong Jin}{yyy,sch}
\end{icmlauthorlist}

\icmlaffiliation{yyy}{Alibaba Group}
%\icmlaffiliation{comp}{Company Name, Location, Country}
\icmlaffiliation{sch}{The author now works at Meta Platforms, Inc}

\icmlcorrespondingauthor{Tian Zhou}{tian.zt@alibaba-inc.com}

\icmlcorrespondingauthor{Rong Jin}{rongjinemail@gmail.com}

\icmlkeywords{Machine Learning, ICML}

\vskip 0.3in
]



\printAffiliationsAndNotice{\icmlEqualContribution} % otherwise use the standard text.

\begin{abstract}
Time series forecasting is essential for many practical applications, with the adoption of transformer-based models on the rise due to their impressive performance in NLP and CV. Transformers' key feature, the attention mechanism, dynamically fusing embeddings to enhance data representation, often relegating attention weights to a byproduct role. Yet, time series data, characterized by noise and non-stationarity, poses significant forecasting challenges. Our approach elevates attention weights as the primary representation for time series, capitalizing on the temporal relationships among data points to improve forecasting accuracy. Our study shows that an attention map, structured using global landmarks and local windows, acts as a robust kernel representation for data points, withstanding noise and shifts in distribution. Our method outperforms state-of-the-art models, reducing mean squared error (MSE) in multivariate time series forecasting by a notable 3.6\% without altering the core neural network architecture. It serves as a versatile component that can readily replace recent patching based embedding schemes in transformer-based models, boosting their performance. The source code for our work is available at: \url{https://anonymous.4open.science/r/AttnEmbed-7430}.

%Our proposed approach surpasses current state-of-the-art models, achieving a significant reduction in mean squared error (MSE) for multivariate time series forecasting by 3.6\%, respectively. Furthermore, our method can be seamlessly integrated as a general component in any existing transformer-based model to replace patching and enhance overall performance. 
\end{abstract}

\section{Introduction}

Time series forecasting is a vital problem that has played an important role in many real-world applications~\cite{wen2022robust, courty1999timing, bose2017probabilistic, li2019enhancing}, ranging from energy, weather, traffic to economics. 
In recent years, traditional statistical and machine learning methods~\cite{arima_1, arima_2} have been gradually replaced by  deep learning models in time series forecasting. 
In particular, CNN and MLP-based models~\cite{timesnet, dlinear} have shown great performance improvement in time series analysis. Moreover, following the successes in NLP \cite{vaswani2017attention, Bert/NAACL/Jacob, gpt2-2019} and CV \cite{Transformers-for-image-at-scale/iclr/DosovitskiyB0WZ21, bao2022beit}, transformer models~\cite{wen2022transformers, zhou2021informer, wu2021autoformer, zhou2022fedformer, Patchformer, liu2023itransformer, card2023} have demonstrated impressive results.
Among the transformer models, PatchTST~\cite{Patchformer} successfully applies the idea of vision transformer (ViT)~\cite{Transformers-for-image-at-scale/iclr/DosovitskiyB0WZ21} to time series by segmenting the time series into multiple patches to serve as input tokens for transformers. While segmentation is beneficial for reducing information redundancy, it overlooks the relationship between a time point and its neighbors, making it insufficient for noise reduction and to handle rapid distribution drifts.

The attention mechanism is a pivotal component that underpins the transformative success of the transformer model across various domains. It is widely regarded as the linchpin behind monumental advancements such as ChatGPT and Midjourney, although other elements like feed-forward networks (FFNs) and positional embeddings also play significant roles. Essentially, the attention mechanism functions as a dynamic, weighted feed-forward layer. Within a self-attention layer, for instance, queries (Q) and keys (K) are used to calculate an attention matrix, which subsequently serves as a weighting matrix that synthesizes the values (V). The resultant attention matrix is usually viewed as a byproduct to reveal the quantitative influence of each input token and to effectively aggregate information across different tokens.

Although time series forecasting can be naturally viewed as a sequence modeling problem, it differs significantly from token sequences in CV or NLP in which limited information can be founded in patches because every data point in time series is simply a scalar. In contrast, tokens in both CV and NLP encompass significantly richer information in that we often find considerable amount of redundant information across different tokens, evidenced by high masking rates used in self supervised learning in CV and NLP. Furthermore, many time series data often contain noises and distribution shifts, partly due to high sampling rates~\cite{wen2022robust}, making the forecasting more challenging. 
These observations inspire us to develop a richer and robust representation for time series data. Since weights in the attention matrix reveal the pairwise relationship between different patches in time series, motivated by the theory of kernel learning~\cite{Wilson2015DeepKL} and reproduced kernel Hilbert space~\cite{Ghojogh2021ReproducingKH}, we propose a novel and robust data representation based on the attention matrix that captures the relationship among different data points in the same time series. One obvious advantage of using attention weights for data representation is that it helps capture the overall seasonality of time series, a special complex relationship. In Section~\ref{sec:provement} and Appendix~\ref{app:theoretical}, we demonstrate that, based on kernel learning theory, employing attention weights as representations more effectively captures the intricate relationships among data points.

%The realm of time series analysis diverges from this notion, not only because an individual data point is merely a scalar value but also because the scalar itself holds limited prognostic significance, as illustrated in Figure \ref{fig:illustration}. The critical factor in this domain is the interplay between data points, which can exhibit highly non-linear relationships.

We also note previous efforts that connect attention mechanism with kernel function. For instance, several studies~\cite{tsai2019transformer, katharopoulos_et_al_2020, song2021implicit} have explored attention from the perspective of kernel functions, either to propose a new paradigm for transformers or to reduce the computational complexity. 
%% This is because attention and kernel functions both calculate the similarity between tokens without being influenced by their order. 
In addition, ~\cite{NIPS1998_226d1f15} exploited non-linear kernel functions to reduce noise in time series while preserving the relationship between different time points. In this study, we also show that using kernel functions, such as polynomial kernels, in attention matrix computation can be more effective for time series forecasting than the standard softmax. 

Several studies have leveraged the attention matrix's pairwise relationships for time series anomaly detection. For example, Anomaly Transformer~\cite{xu2021anomaly} introduces an association discrepancy by measuring the Kullback-Leibler divergence between the attention matrix and a learnable Gaussian kernel to identify anomalies. Similarly, DCdetector~\cite{Yang2023DCdetectorDA} suggests that the divergence between attention matrices is a dependable indicator. 
However, our work is not limited by the anomaly detection framework. Instead, we have developed a generalized data representation from the attention matrix, which presents versatile potential for tasks involving embeddings.

%This suggests that the attention map can be more than a means to an end; it can be an endpoint in itself. In this study, we delve deeply into the notion that the attention map alone can be a robust alternative to patch-based information density approaches, possessing substantial representational strength independently.

%Therefore, attention weight is a powerful representation that can effectively remove noise while retaining maximum information. We also show that kernel functions, such as RBF and polynomial kernels, can be beneficial for representing time series data.

Our contributions in  this paper are summarized as follows:

\begin{itemize}
    \item Attention as robust representation: We propose a novel time series representation method called AttnEmbed, which utilizes attention weights as representation of time segments. The resilience of AttnEmbed to both noise and non-stationary distributions is verified by our empirical studies of synthetic datasets, and is also verified by our theoretical analysis.  

    \item Outstanding performance for time series forecasting: 
    Our innovative embedding schema, AttnEmbed, integrates a global landscape and smoothing design to adeptly handle distribution shifts. When paired with a vanilla transformer, this approach significantly outperforms state-of-the-art methods in time series forecasting, as evidenced by our comprehensive experimental analysis.
    
 %   the changes over time.
 %   \item Performance \& Experiments: Our real-world experiments demonstrate that AttnEmbed outperforms current state-of-the-art methods in multivariate time series forecasting by $3.6\%$ reduction on MSE. Moreover, we employ synthetic datasets to illustrate the efficiency of non-stationary processing and noise reduction techniques.
    \item Kernel functions for better attention: We illustrate that the polynomial kernel can effectively replace traditional similarity measures in attention mechanisms, yielding representations that enhance performance in forecasting tasks.
    %The embeddings from kernel functions can also achieve remarkable performance comparable to attention and state-of-the-art methods.
    
    \item General plug-in: AttnEmbed can be seamlessly integrated as a general plug-in module. We have effectively integrated it into multiple methods, yielding performance enhancements over the patching method.

\end{itemize}

\section{Attention as Robust Representation}

To verify the resilience of attention as a data representation to both noise and non-stationary distributions, we first conduct experiments on synthetic data, and then examine the robustness of attention based representation by a theoretical analysis. 
%%Finally, as a side effect of robust representation,  we will show that the proposed attention based representation can effectively reduce the risk of rank collapse with transformer.

\subsection{An Empirical Study on Synthetic Data}

We develop two synthetic datasets, one for non-stationary time series and one for noisy data, and compare our approach (i.e. AttnEmbed), against a method that inputs patches of the original data with linear projection for embedding (i.e., PatchTST,VIT). 

\begin{figure}[h]
    \centering
    \includegraphics[width=0.48\textwidth]{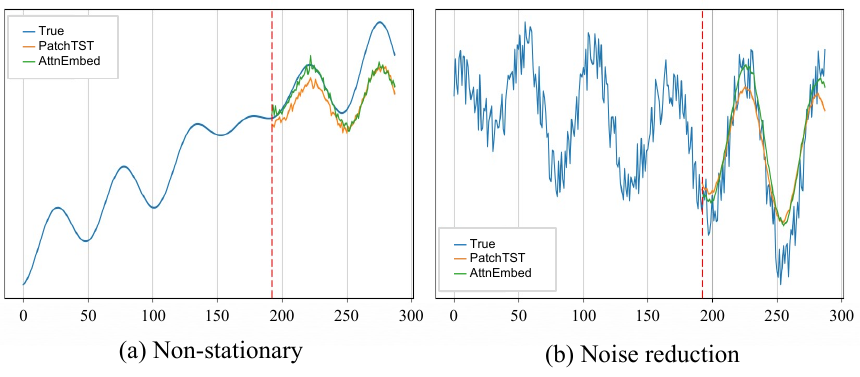}
    \vskip -0.1in
    \caption{Comparison between AttnEmbed (ours) and PatchTST on synthetic data. (a) Non-stationary. (b) Noise reduction.
    % (a) Non-stationary data generated by $f_1(\cdot)$. (b) Synthetic data with noise generated by $f_2(\cdot)$. The red dot line marks the beginning of forecasting.
    }
    \label{fig:synthetic}
    \vskip -0.15in
\end{figure}

\paragraph{Synthetic Data.} The synthetic data is generated by the aggregation of 10 sinusoids and cubic functions, each characterized by distinct random parameters:
\begin{align*}
f_1(x) &=& \mkern-18mu \sum Asin(\omega x + \phi) &+ \sum (ax^3 + bx^2 + cx + d), \\
f_2(x) &=& \mkern-18mu \sum Asin(\omega x + \phi) &+ \sum (ax^3 + bx^2 + cx + d) + \sigma,
\end{align*}
%\begin{equation}
%    f_1(x) = \sum Asin(\omega x + \phi) + \sum (ax^3 + bx^2 + cx + d),
%\end{equation}
%\begin{equation}
%    f_2(x) = \sum Asin(\omega x + \phi) + \sum (ax^3 + bx^2 + cx + d) + \sigma,
%\end{equation}
where $f_1(x)$ is a designed for non-stationary distribution and $f_2(x)$ is designed for noisy data. All the parameters in $f_1(\cdot)$ and $f_2(\cdot)$ are randomly chosen. 
A total of 2000 time steps are sampled, with a lookback window size of 192 and the forecast horizon of 96.
Figure \ref{fig:synthetic} shows the  plots of the two functions, together with all experimental results for comparison.

\paragraph{Non-stationary Data.} Figure \ref{fig:synthetic} (a) shows 
time series data generated by $f_1(\cdot)$, which clearly show a noticeable shift. We can observe that the proposed representation AttnEmbed is able to better capture the overall drift than PatchTST for the first 50 time points of forecasting, and the advantage disappears after time point $250$.

%PatchTST
%PatchTST fails to capture this evolving trend, leading to forecasted scale that were notably undersized. PatchTST does not take into account the relationships within patches, thus discarding the latest distribution changes. However, AttnEmbed adeptly tracks the trend of historical fluctuations in the early half of its forecasting, illustrating that the inter-window attention weights are capable of utilizing mutual information to capture information regarding distribution shifts.

\paragraph{Noise Reduction.} Figure~\ref{fig:synthetic} (b) shows the time series generated by $f_2(\cdot)$ (i.e. $f_1(\cdot)$ plus noise). Experimental results indicate that while both AttnEmbed and PatchTST effectively mitigate noise, AttnEmbed delivers more accurate predictions for imminent time points in forecasting.

In conclusion, our empirical studies using synthetic data demonstrate that attention-based embedding is an effective schema for addressing noise and non-stationary distributions.
%excels by preserving a substantial amount of information, which facilitates the accurate forecasting of future amplitudes while simultaneously reduce noise. Conversely, PatchTST fails in achieving this delicate balance.

\subsection{Theoretical Analysis for Robustness of Attention based Representation}
\label{sec:provement}

To demonstrate that AttnEmbed is more resilient to noise, we first will show that by adding significant amount of noises to the input patterns, the distance of ``similar'' data pairs can be very close to that for ``dissimilar'' pairs. In constrast, by using attention based representations, we are able to maintain that the distance for ``similar'' data pairs is significantly smaller than that for ``dissimilar'' pairs even after adding large noises to the input patterns. Below, we will provide the sketch of overall results, and postpone the full analysis to the appendix. 

Consider we have $n$ vectors $x_i \in \mathbb{R}^d$ in a sequence that are generated from $m < d$ Gaussian distributions $\mathcal{N}(\mu_i, I_d), i=1, \ldots, m$. We assume $\langle \mu_i, \mu_j \rangle = \delta_{i,j}s$. It is easy to show that for two ``similar'' data points $x_i^+$ and $x_j^+$ that are generated from the same distribution, their expected distance is $\E[|x_i^+ - x_j^+|^2] = 2d$, whereas for two ``dissimilar'' data points $x_i^-$ and $x_j^+$ that are generated from different distributions, their expected distance is $\E[|x_i^- - x_j^-|^2] = 2d + 2s$. When $s \ll d$, i.e., noises are much larger than signals, we have $\E[|x_i^+ - x_j^+|^2] \approx \E[|x_i^- - x_j^-|^2]$, implying that there is a significant chance that $|x_i^- - x_j^-|^2$ can be noticeably smaller than $|x_i^+ - x_j^+|^2$. Now, if we use the attention weights as the representation, denoted by $f(x)$, using the same notation for ``similar'' and ``dissimilar'' data pairs, we can show that
\[
\frac{\E[|f(x_i^-) - f(x_j^-)|^2] - \E[|f(x_i^+) - f(x_j^+)|^2]}{|f(x_i^-) - f(x_j^-)|^2]} = \Omega(1)
\]
with appropriate choice of temperature. It implies that even after adding large noises to input patters, we can still clearly distinguish ``similar'' data pairs from the ``dissimilar'' data pairs, thus verifying the robustness of the proposed attention based data representation. The full theoretical analysis is in appendix ~\ref{app:theoretical}.

\section{Related Work}

In this section, we provide brief reviews of literature in the areas of  time series forecasting and the relationship between attention mechanism and kernel function.

\subsection{Time Series Forecasting}

Recently, inspired by great success in NLP and CV, transformer models have also been widely used in time series forecasting \cite{wen2022transformers}. 
% Here, we summarize several representative algorithms. 
Informer \cite{zhou2021informer} proposes a probability sparse attention mechanism to deal with long-term dependencies. 
Autoformer \cite{wu2021autoformer} introduces a decomposition transformer architecture and replaces the attention module with an Auto-Correlation mechanism. 
FEDformer \cite{zhou2022fedformer} employs a Fourier-enhanced architecture to improve computational efficiency, achieving linear complexity.
PatchTST \cite{Patchformer} segments time series into individual patches, which successfully increases input length and reduce information redundancy.
GPT4TS \cite{zhou2023ofa} utilizes a frozen GPT-2 and achieves a promising performance in several time series tasks.
CARD \cite{card2023} and iTransformer \cite{liu2023itransformer} integrates the correlations among multiple variables to enhance the performance in multivariate time series forecasting.
Moreover, TimesNet \cite{timesnet} treats time series as a 2D signal and utilizes a convolution-based inception network as its backbone.
A simple MLP-based DLinear \cite{dlinear} outperforms a lot of transformer models in time series forecasting with channel-independence and seasonal-trend decomposition.

\subsection{Attention and Kernel Function}

The perspective of viewing attention as a kernel function is widely recognized in the literature, encompassing modifications to transformers and attention mechanisms~\cite{tsai2019transformer, song2021implicit}, acceleration of computational processes~\cite{katharopoulos_et_al_2020}, and time series anomaly detection~\cite{Yang2023DCdetectorDA, xu2021anomaly}.
\cite{tsai2019transformer} proposes that the attention mechanisms in transformers can be interpreted as employing a kernel smoother across the input data and the kernel scores is the similarities between inputs. \cite{song2021implicit} derives that the attention is a product of RBF kernel and the exponential of $\ell_2$-norm. Also, given kernel functions are advantageous in computational efficiency for distance calculations, \cite{katharopoulos_et_al_2020} reformulates self-attention as a linear operation involving the dot-product of kernelized feature maps. 
Excitingly, the exploration of kernel functions has extended into the realm of time series anomaly detection. 
Anomaly Transformer~\cite{xu2021anomaly} and DCDetector~\cite{Yang2023DCdetectorDA} both utilize Kullback-Leibler divergence to calculate the distance between attention matrix and Gaussian kernel, establishing a novel linkage between attention mechanisms and kernel functions in the domain of time series.
Although the methods mentioned previously utilize attention weights primarily for token mixing, our work is among the first to explore attention as an end in itself—not just a means—for embedding schema in the field of time series forecasting. To our knowledge, such an approach has been rarely investigated.

\section{Methodology}
\begin{figure*}[h]
    \centering
    \includegraphics[width=0.88\textwidth]{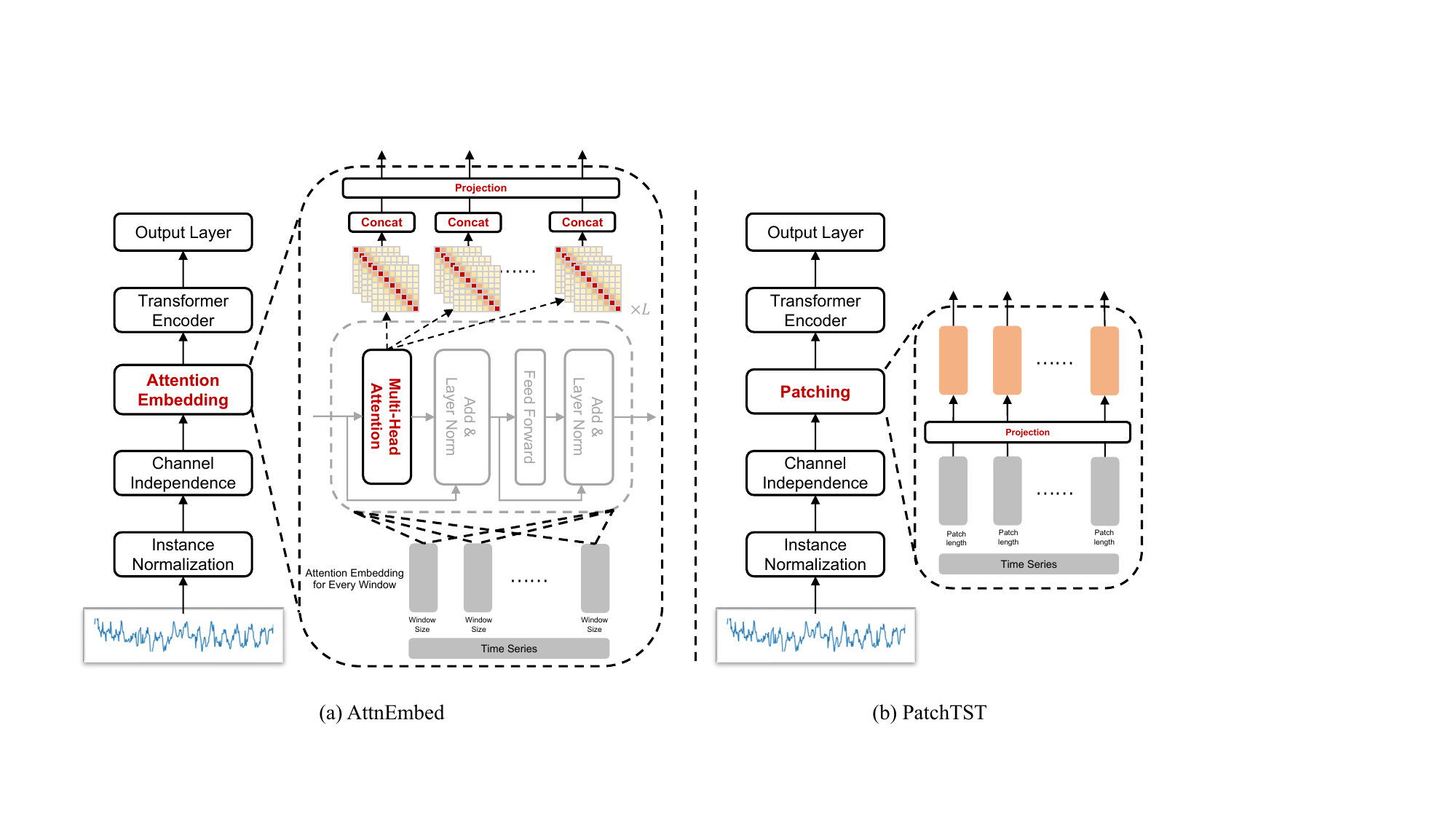}
    \vskip -0.15in
    \caption{The architecture of (a) AttnEmbed and a comparison with (b) PatchTST. Unlike PatchTST, AttnEmbed considers the relationship of time steps within each window.}
    \label{fig:model_structure}
    \vskip -0.15in
\end{figure*}

Consider a multivariate time series with look back window $L$: $(x_1, ..., x_t, ..., x_L)$, where $x_t \in \mathbb{R}^M$ is the observation at time $t$ with $M$ channels. Our objective is to forecast future steps with a horizon of $T$, denoted by $(x_{L+1}, ..., x_{L+T})$.

\subsection{Overall Architecture}

The architecture of AttnEmbed is illustrated in Figure \ref{fig:model_structure}.
% 关于attnembed的介绍
AttnEmbed contains Pre-processing module which consists of instance normalization and channel independence, Attention Embedding module and Transformer Encoder.It is important to note that our proposed method serves as a model-agnostic alternative for embedding, with the transformer employed merely as an illustrative example. As demonstrated in Table \ref{tab:plugin}, we have conducted experiments with various baseline models, including PatchTST~\cite{Patchformer} and CARD~\cite{card2023}.
\paragraph{Pre-process Module.} The input time series in Pre-process module is first normalized by instance normalization~\cite{kim2022reversible}. This normalization block performs a simple normalization of the input time series with mean and variance, and subsequently integrates these values back to the output. We then employ the channel independence technique, as used in DLinear \cite{dlinear} and PatchTST \cite{Patchformer}, which has been widely validated for its effectiveness in time series forecasting. This technique essentially transforms a multivariate time series forecasting problem into a univariate one.
\paragraph{Attention Embedding Module.} The Attention Embedding module is critical in the architecture of AttnEmbed. The pre-processed time series is split into multiple windows in the Attention Embedding module. Within each window, we utilize a shared Embedding self-attention block with $L$ layers to extract the mutual relationships between time steps. Specifically, for each window, we extract the intermediary computational outputs generated by the Embedding module, obtaining a set of attention matrices. Then, all the last row of attention matrices are concatenated to form the embedding for the respective window.
\paragraph{Transformer Encoder.} Similar to PatchTST~\cite{Patchformer},  next we employ a Transformer Encoder based on the generated embeddings for the forecasting task.
% attnembed和patchtst的区别
As shown in Figure \ref{fig:model_structure}, compared to PatchTST, the primary distinction is that AttnEmbed integrates the interaction between time steps within a single window or patch, which is essential for addressing distribution shift by capturing local dynamics.
%Once all the embeddings have been generated, they can be applied to downstream tasks. Similar to PatchTST \cite{Patchformer},  we employ a Transformer Encoder for the forecasting task.
% attnembed和patchtst的区别
%As shown in Figure \ref{fig:model_structure}, compared to PatchTST, the primary distinction is that AttnEmbed integrates the interaction between time steps within a single window or patch, which is critical for capturing local information.

\subsection{Attention Embedding}
\label{attn_emb}
We now delve into the specifics of computing the attention embedding, as depicted in Figure \ref{fig:attn_embed}. The pre-processed univariate time series is represented as $U = [u_1, ..., u_t, ..., u_L] \in \mathbb{R}^{L}$, where $L$ is the length of the series. 

\begin{figure}[h]
    \centering
    \includegraphics[width=0.4\textwidth]{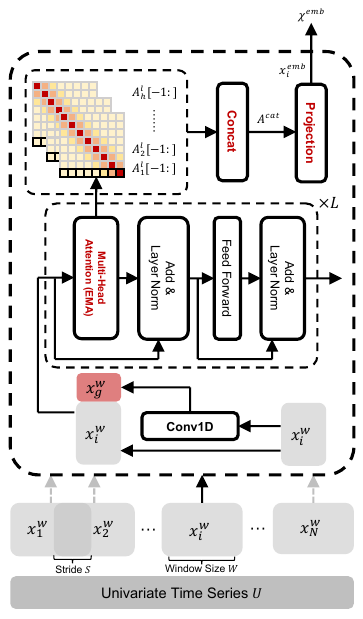}
    \vskip -0.1in
    \caption{Detail of attention embedding. 
    % The pre-processed univariate time series is first split into several windows. Within each window, a Conv1D is employed to generate global landmarks. Subsequently, transformer augmented with EMA is utilized to generate attention matrices. The final rows from these matrices are concatenated and fed into a projection layer to complete the embedding.
    }
    \label{fig:attn_embed}
    \vskip -0.15in
\end{figure}

\subsubsection{Tokenization and Global Landmark}

Each input univariate time series is split into several overlapped or non-overlapped windows with window size $W$ and stride length $S$. Thus, the raw tokens are generated as $\mathcal{\hat{X}} = [x^w_1, ..., x^w_i, ..., x^w_N] \in \mathbb{R}^{W \times 1 \times N}$, where $N = \lfloor \frac{L - W}{S} \rfloor + 1$. Each window, comprising $W$ time steps, is processed through a common set of self-attention layers to yield a concatenated attention matrix, which is then utilized as the embedding.

While the above attention embedding method benefits from capturing local information, it overlooks the global information of the time series. Thus, we introduce global landmarks designed to incorporate the information from the entire series. We utilze Conv1D to calculate the global landmarks:
\begin{equation}
    \label{global_landmark}
    x^w_g = {\rm Conv1D}(x^w_i),
\end{equation}
where $x^w_g \in \mathbb{R}^{G}$ and $G$ represents the number of landmarks, which is dictated by the parameters of Conv1D. Subsequently, the embedding matrix formed by the shared attention layers is assembled by concatenating each local feature representation $x^w_t$ with the corresponding global feature representation $x^w_g$:

\begin{equation}
    \label{input_token}
    \mathcal{X} = [[x^w_g, x^w_1], ..., [x^w_g, x^w_N]],
\end{equation}
where $\mathcal{X} \in \mathbb{R}^{(G + W) \times 1 \times N}$.

For each individual window, the attention score of the $h$-th head in the $l$-th layer is denoted as $A^l_h \in \mathbb{R}^{(G + W) \times (G + W)}$
% , which contains the mutual relationships between different time points. 
Through the combination and projection of these attentions, an embedding can be generated that characterizes the local information of the window. Specifically, the final rows from all attention matrices are concatenated and subsequently passed through a projection layer:
\begin{equation}
    \label{projection_embedding}
    x^{emb} = {\rm Proj}(A^{cat}),
\end{equation}
\begin{equation}
    \label{concat_attention}
    A^{cat} = \mathop{{\rm Concat}}\limits_{L^a in [1, L^a], h \in [1, H^a]]}(A^l_h[-1, :]),
\end{equation}
where $A^{cat} \in \mathbb{R}^{L^a H^a (G + W)}$, $L^a$ and $H^a$ represent the number of layers and the number of heads in the set of self-attention layers of the embedding module respectively.
Thus the final output embedding can be denoted as $\mathcal{X}^{emb} = [x^{emb}_1, ..., x^{emb}_N]$.

\subsubsection{Exponential Moving Average (EMA)}

To enhance the capture of local information, we have incorporated an Exponential Moving Average (EMA) within the self-attention blocks of the embedding module. EMA is a special case of moving average that responds to changes more quickly in time and can smooth out the output for noise reduction. Specifically, EMA utilizes factors exponentially decaying weighting factors as:
\begin{equation}
    \label{ema}
    y_t = \alpha x_t + (1 - \alpha) y_{t-1},
\end{equation}
where $\alpha \in (0, 1)$ is the degree of weighting decrease. Many works \cite{ma2023mega, card2023} have explored the application of EMA in the attention module. We integrate EMA into the queries and keys within the attention mechanism, opting for a non-parametric approach to reinforce stability during the training process.

\subsection{Kernel Function for Attention based Representation}
\label{sec:kernel_function}
Inspired by earlier studies \cite{choromanski2021rethinking, katharopoulos_et_al_2020} that pioneered a kernel-based interpretation of the attention matrix, we propose the adoption of advanced kernel methods. We utilize both the Radial Basis Function (RBF) and polynomial kernels to assess the degree of similarity between time steps within a given window. This methodological innovation underpins the output of our attention embedding module, thereby replacing the $\mathcal{X}^{emb}$ as described in Section \ref{attn_emb}.

We generate queries $Q$ and keys $K$ by linearly projecting the token tensor $\mathcal{X}_i = [x^w_g, x^w_i]^T \in \mathbb{R}^{(G + W) \times 1}$ as follows:
\begin{equation}
Q = F_q(\mathcal{X}i), \quad K = F_k(\mathcal{X}i),
\end{equation}
with both $Q, K \in \mathbb{R}^{(G + W) \times d}$, where $F_q, F_k$ map from dimension $1$ to $d$ through MLP layers. These matrices are further processed to obtain $Q_h, K_h \in \mathbb{R}^{(G + W) \times d{head}}$, corresponding to the queries and keys for the $h^{th}$ attention head, with $d = H_a \times d{head}$.

Kernel-based embeddings are then computed as:
\begin{align}
    x^{emb}_{kernel} &= Proj(A^{cat}_{kernel}), \\
    A^{cat}_{kernel} &= \mathop{{\rm Concat}}\limits_{h \in [1, H^a]}\mathcal{K}(Q_h[-1, :], K_h),
\end{align}
where $\mathcal{K}$ is the kernel function. In this paper, we introduce two kernel functions, the RBF kernel and the polynomial kernel.
% :
% \begin{align}
%     {\rm RBF: }& \mathcal{K}_{rbf}(x, y) = {\rm exp}(-\frac{\Vert x - y \Vert^2}{2\sigma^2}), \\
%     {\rm Polynomial: }& \mathcal{K}_{poly}(x, y) = (x^Ty + c)^p,
% \end{align}
% where $\sigma$ is a learnable parameter of RBF kernel, $c \geq 0$ is the constant offset added to inner product and $p$ is the degree of polynomials in polynomial kernel.

\section{Experiments}

\subsection{Experiments on Real-world Datasets}

\paragraph{Datasets.} We conduct experiments on seven popular real-world benchmark datasets, including 4 ETT dataset~\cite{zhou2021informer} (comprising of two hourly dayasets ETTh1, ETTh2 and two 15-minute datasets ETTm1, ETTm2), 
the Electricity \footnote{https://archive.ics.uci.edu/ml/datasets/ElectricityLoadDiagrams 20112014} dataset for hourly electricity consumption, 
the Weather \footnote{https://www.bgc-jena.mpg.de/wetter/} dataset for 10-minute weather forecasting and 
the Traffic \footnote{http://pems.dot.ca.gov} dataset for hourly road occupancy rate.
\paragraph{Baselines.}
In our comparison, we have chosen a range of representative baselines, including transformer-based models such as PatchTST \cite{Patchformer}, FiLM \cite{NEURIPS2022_film}, FEDformer \cite{zhou2022fedformer}, Autoformer \cite{wu2021autoformer}, and Informer \cite{zhou2021informer}; the MLP-based DLinear \cite{dlinear}; and the CNN-based TimesNet \cite{timesnet}. Our research is particularly concerned with exploring interactions within individual channels, so we've limited our benchmarking to cutting-edge models that adopt a channel-independent structure. This selection criterion ensures a focused and pertinent benchmarking against our research aims. Consequently, models like iTransformer \cite{liu2023itransformer} and CARD \cite{card2023} are excluded from the primary experiments. Nonetheless, in Section \ref{plugin}, we demonstrate how AttnEmbed can be effectively integrated as a plug-in to enhance CARD's performance.

\paragraph{Main Results.}
For better comparison, we follow the experimental settings in \cite{timesnet}, maintaining the lookback length at 96, and the horizon length at 96, 192, 336, and 720, respectively. The main results of multivariate forecasting are summarized in Table \ref{tab:multi_forecasting}. The lower MSE/MAE indicates the better forecasting results. AttnEmbed notably achieves state-of-the-art results, outshining the top-performing PatchTST model. Crucially, this is achieved while preserving a similar main model architecture, specifically the vanilla transformer encoder used by PatchTST. The enhancement in performance can be solely attributed to our shift from traditional patching to the AttnEmbed method, marking this advancement as substantial. AttnEmbed gains the best performance on 6 out of 7 datasets in both MSE and MAE. Compared with PatchTST, AttnEmbed yield an overall 3.6\% relative MSE reduction and 2.1\% relative MAE reduction. In datasets with noisier and more frequently shifting distributions, such as ETTh1 and Traffic, the improvement from PatchTST to AttnEmbed is more pronounced, achieving a significant reduction in MSE by 6.2\% and 8.2\%, respectively. In general, the improvements made by AttnEmbed are consistent across various horizons, indicating that attention effectively represents time series with the resilience of both noise and non-stationary distributions. 

\begin{table*}[h]
    \vskip -0.10in
    \caption{Multivariate forecasting with a lookback length of 96. All models are averaged from 4 different horizons. A lower MSE indicates better performance. The best ones are in Bold, and the second ones are \underline{underlined}. Detailed results are provided in Appendix \ref{app:full_multi}}
    \label{tab:multi_forecasting}
    \begin{center}
    \scalebox{0.82}{
    \small
    \begin{tabular}{c|c|cc|cc|cc|cc|cc|cc|cc|cc}
    \toprule
    \midrule
    
    \multicolumn{2}{c|}{Methods}&\multicolumn{2}{c|}{\bf{AttnEmbed}}&\multicolumn{2}{c|}{PatchTST}&\multicolumn{2}{c|}{TimesNet}&\multicolumn{2}{c|}{DLinear}&\multicolumn{2}{c|}{FiLM}&\multicolumn{2}{c|}{FEDformer}&\multicolumn{2}{c|}{Autoformer}&\multicolumn{2}{c}{Informer} \\
    
    \midrule
    
    \multicolumn{2}{c|}{Metric} & MSE& MAE & MSE & MAE& MSE & MAE& MSE  & MAE& MSE  & MAE& MSE  & MAE& MSE  & MAE & MSE  & MAE \\
    \midrule
    
    % \multirow{5}{*}{\rotatebox{90}{$Weather$}}
    % & 96  &\bf{0.171}&\bf{0.215}&0.178&\underline{0.219}&\underline{0.172}&0.220&0.196&0.255&0.193&0.234&0.217&0.296&0.266&0.336&0.300&0.384\\
    % & 192 &\bf{0.218}&\bf{0.257}&0.224&\underline{0.259}&\underline{0.219}&0.261&0.237&0.296&0.236&0.269&0.276&0.336&0.307&0.367&0.598&0.544\\
    % & 336 &\bf{0.274}&\bf{0.297}&\underline{0.278}&\underline{0.298}&0.280&0.306&0.283&0.335&0.288&0.304&0.339&0.380&0.359&0.395&0.578&0.523\\
    % & 720 &\underline{0.348}&\bf{0.346}&0.350&\bf{0.346}&0.365&0.359&\bf{0.345}&0.381&0.358&0.350&0.403&0.428&0.419&0.428&1.059&0.741\\
    % & Avg 
    \multicolumn{2}{c|}{Weather}&\bf{0.252}&\bf{0.278}&\underline{0.257}&\underline{0.280}&0.259&0.287&0.265&0.317&0.269&0.339&0.309&0.360&0.338&0.382&0.634&0.548\\
    \midrule
    
    % \multirow{5}{*}{\rotatebox{90}{$ETTh1$}}
    % & 96  &\bf{0.367}&\bf{0.398}&0.393&0.408&0.384&0.402&0.386&\underline{0.400}&0.388&0.401&\underline{0.376}&0.419&0.449&0.459&0.865&0.713\\
    % & 192 &\bf{0.420}&\bf{0.428}&0.445&0.434&\underline{0.436}&\underline{0.429}&0.437&0.432&0.443&0.439&\bf{0.420}&0.448&0.500&0.482&1.008&0.792\\
    % & 336 &\bf{0.448}&\bf{0.438}&0.484&\underline{0.451}&0.491&0.469&0.481&\underline{0.459}&0.488&0.466&0.459&0.465&0.521&0.496&1.107&0.809\\
    % & 720 &\bf{0.454}&\bf{0.459}&\underline{0.480}&\underline{0.471}&0.521&0.500&0.519&0.516&0.525&0.519&0.506&0.507&0.514&0.512&1.181&0.865\\
    % & Avg 
    \multicolumn{2}{c|}{ETTh1}&\bf{0.422}&\bf{0.430}&0.450&\underline{0.440}&0.458&0.450&0.456&0.452&0.461&0.456&\underline{0.440}&0.460&0.496&0.487&1.040&0.795\\
    \midrule
    
    % \multirow{5}{*}{\rotatebox{90}{$ETTh2$}}
    % & 96  &\underline{0.296}&\underline{0.346}&\bf{0.294}&\bf{0.343}&0.340&0.374&0.333&0.387&0.296&0.344&0.358&0.397&0.346&0.388&3.755&1.525\\
    % & 192 &\bf{0.369}&\bf{0.392}&\underline{0.377}&\underline{0.393}&0.402&0.414&0.477&0.476&0.389&0.402&0.429&0.439&0.456&0.452&5.602&1.931\\
    % & 336 &\bf{0.376}&\bf{0.405}&\underline{0.381}&\underline{0.409}&0.452&0.452&0.594&0.541&0.418&0.430&0.496&0.487&0.482&0.486&4.721&1.835\\
    % & 720 &\bf{0.405}&\bf{0.432}&\underline{0.412}&\underline{0.471}&0.462&0.468&0.831&0.657&0.433&0.448&0.463&0.474&0.515&0.511&3.647&1.625\\
    % & Avg 
    \multicolumn{2}{c|}{ETTh2}&\bf{0.361}&\bf{0.393}&\underline{0.366}&\underline{0.404}&0.414&0.427&0.559&0.515&0.384&0.406&0.437&0.449&0.450&0.459&4.431&1.729\\
    \midrule
    
    % \multirow{5}{*}{\rotatebox{90}{$ETTm1$}}
    % & 96  &\bf{0.317}&\bf{0.356}&\underline{0.321}&\underline{0.360}&0.338&0.375&0.345&0.372&0.348&0.367&0.379&0.416&0.505&0.475&0.672&0.571\\
    % & 192 &\bf{0.357}&\bf{0.381}&\underline{0.362}&\underline{0.384}&0.371&0.387&0.380&0.389&0.387&0.385&0.426&0.441&0.553&0.496&0.795&0.669\\
    % & 336 &\bf{0.387}&\underline{0.404}&\underline{0.392}&\bf{0.402}&0.410&0.411&0.413&0.413&0.418&0.405&0.445&0.459&0.621&0.537&1.212&0.871\\
    % & 720 &\bf{0.448}&\underline{0.439}&\underline{0.450}&\bf{0.435}&0.478&0.450&0.474&0.453&0.479&0.440&0.543&0.490&0.671&0.561&1.166&0.823\\
    % & Avg 
    \multicolumn{2}{c|}{ETTm1}&\bf{0.377}&\bf{0.395}&\underline{0.381}&\bf{0.395}&0.400&0.406&0.403&0.407&0.408&0.399&0.448&0.452&0.588&0.517&0.961&0.734\\
    \midrule
    
    % \multirow{5}{*}{\rotatebox{90}{$ETTm2$}}
    % & 96  &\underline{0.181}&\underline{0.265}&\bf{0.178}&\bf{0.260}&0.187&0.267&0.193&0.292&0.183&0.266&0.203&0.287&0.255&0.339&0.365&0.453\\
    % & 192 &\bf{0.245}&\bf{0.304}&0.249&0.307&0.249&0.309&0.284&0.362&\underline{0.247}&\underline{0.305}&0.269&0.328&0.281&0.340&0.533&0.563\\
    % & 336 &\bf{0.309}&0.349&0.313&\underline{0.346}&0.321&0.351&0.369&0.427&\bf{0.309}&\bf{0.343}&0.325&0.366&0.339&0.372&1.363&0.887\\
    % & 720 &0.409&0.407&\bf{0.400}&\bf{0.398}&0.408&0.403&0.554&0.522&\underline{0.407}&\bf{0.398}&0.421&0.415&0.433&0.432&3.379&1.338\\
    % & Avg 
    \multicolumn{2}{c|}{ETTm2}&\underline{0.286}&0.331&\bf{0.285}&\bf{0.327}&0.291&0.333&0.350&0.401&0.287&\underline{0.328}&0.305&0.349&0.327&0.371&1.410&0.810\\
    \midrule
    
    % \multirow{5}{*}{\rotatebox{90}{$ECL$}}
    % & 96  &\bf{0.166}&\bf{0.252}&0.174&\underline{0.259}&\underline{0.168}&0.272&0.197&0.282&0.198&0.276&0.193&0.308&0.201&0.317&0.274&0.368\\
    % & 192 &\bf{0.172}&\bf{0.259}&\underline{0.178}&\underline{0.265}&0.184&0.289&0.196&0.285&0.198&0.279&0.201&0.315&0.222&0.334&0.296&0.386\\
    % & 336 &\bf{0.191}&\bf{0.277}&\underline{0.196}&\underline{0.282}&0.198&0.300&0.209&0.301&0.217&0.301&0.214&0.329&0.254&0.361&0.300&0.394\\
    % & 720 &\underline{0.231}&\bf{0.309}&0.237&\underline{0.316}&\bf{0.220}&0.320&0.245&0.333&0.279&0.357&0.246&0.355&0.254&0.361&0.373&0.439\\
    % & Avg 
    \multicolumn{2}{c|}{ECL}&\bf{0.189}&\bf{0.274}&0.196&\underline{0.280}&\underline{0.192}&0.295&0.212&0.300&0.223&0.303&0.214&0.327&0.227&0.338&0.311&0.397\\
    \midrule
    
    % \multirow{5}{*}{\rotatebox{90}{$Traffic$}}
    % & 96  &\bf{0.428}&\bf{0.276}&\underline{0.477}&\underline{0.305}&0.593&0.321&0.650&0.396&0.649&0.391&0.587&0.366&0.613&0.388&0.274&0.368\\
    % & 192 &\bf{0.434}&\bf{0.274}&\underline{0.471}&\underline{0.299}&0.617&0.336&0.598&0.370&0.603&0.366&0.604&0.373&0.616&0.382&0.296&0.386\\
    % & 336 &\bf{0.448}&\bf{0.282}&\underline{0.485}&\underline{0.305}&0.629&0.336&0.605&0.373&0.613&0.371&0.621&0.383&0.622&0.337&0.300&0.394\\
    % & 720 &\bf{0.478}&\bf{0.299}&\underline{0.518}&\underline{0.325}&0.640&0.350&0.645&0.394&0.692&0.427&0.626&0.382&0.660&0.408&0.373&0.439\\
    % & Avg 
    \multicolumn{2}{c|}{Traffic}&\bf{0.447}&\bf{0.282}&\underline{0.487}&\underline{0.308}&0.620&0.336&0.625&0.383&0.639&0.389&0.610&0.376&0.628&0.379&0.311&0.397\\
    \midrule
    
    \bottomrule
    \end{tabular}
    }
    \end{center}
    \vskip -0.15in
\end{table*}
Recent works have shown that extending the lookback length can enhance performance. Thus, we have also demonstrated that AttnEmbed's effectiveness is not constrained by the lookback window size and can outperform PatchTST with longer inputs, in Section \ref{plugin}.

\subsubsection{Kernel Functions}
Our experiments with real-world datasets (ETTh1, ETTm1, and Weather) using RBF and polynomial kernels demonstrate that kernel functions can achieve results comparable to softmax-based attention mechanisms. The summarized outcomes in Table \ref{tab:kernel_forecasting} indicate that both kernels not only meet but, on average, surpass the performance of previous state-of-the-art (SOTA) models like PatchTST and TimesNet in terms of MSE and MAE. Impressively, the polynomial kernel attains a relative MSE reduction of $4.2\%$ and MASE reduction of $2.0\%$ compared to PatchTST on ETTh1. These findings suggest that the effectiveness of attention weights can be replicated through a kernel approach, where similarities between tokens are calculated. Consequently, this validates the integration of kernel functions into time series forecasting, underscoring their viability and promising potential for such applications.

\begin{table*}
\vskip -0.10in
\caption{Multivairate forecasting results with RBF kernel and polynomial kernel with a lookback length of 96. All models are averaged from on 4 different horizons. A lower MSE indicates better performance. The best ones are in Bold, and the second ones are \underline{underlined}. Detailed results are provided in Appendix \ref{app:kernel_full}}
\label{tab:kernel_forecasting}
\begin{center}

\scalebox{0.82}{
\begin{tabular}{c|c|cc|cc|cc|cc|cc}
\toprule
\midrule

\multicolumn{2}{c|}{Methods}&\multicolumn{2}{c|}{AttnEmbed}&\multicolumn{2}{c|}{RBF Kernel}&\multicolumn{2}{c|}{Polynomial Kernel}&\multicolumn{2}{c|}{PatchTST}&\multicolumn{2}{c}{TimesNet} \\

\midrule

\multicolumn{2}{c|}{Metric} & MSE& MAE & MSE & MAE& MSE & MAE& MSE  & MAE& MSE  & MAE\\
\midrule

% \multirow{5}{*}{\rotatebox{90}{$Weather$}}
% & 96  &\bf{0.171}&\bf{0.215}&0.175&0.220&0.174&\underline{0.216}&0.178&0.219&\underline{0.172}&0.220\\
% & 192 &\bf{0.218}&\bf{0.257}&0.222&0.258&0.221&\bf{0.257}&0.224&0.259&\underline{0.219}&0.261\\
% & 336 &\underline{0.274}&\underline{0.297}&0.276&\underline{0.297}&\bf{0.272}&\bf{0.296}&0.278&0.298&0.280&0.306\\
% & 720 &\underline{0.348}&\bf{0.346}&\bf{0.347}&\bf{0.346}&0.350&\bf{0.346}&0.350&\bf{0.346}&0.365&0.359 \\
% & Avg 
\multicolumn{2}{c|}{Weather}&\bf{0.252}&\bf{0.278}&0.255&0.280&\underline{0.254}&\underline{0.279}&0.257&0.280&0.259&0.287 \\
\midrule

% \multirow{5}{*}{\rotatebox{90}{$ETTh1$}}
% & 96  &\bf{0.367}&\bf{0.398}&\underline{0.374}&\bf{0.398}&0.380&0.400&0.393&0.408&0.384&0.402\\
% & 192 &\bf{0.420}&\bf{0.428}&0.441&0.436&0.437&0.431&0.445&0.434&\underline{0.436}&\underline{0.429}\\
% & 336 &\bf{0.448}&\bf{0.438}&0.475&0.452&\underline{0.457}&\underline{0.442}&0.484&\underline{0.451}&0.491&0.469\\
% & 720 &\underline{0.454}&\underline{0.459}&0.491&0.472&\bf{0.450}&\bf{0.453}&0.480&0.471&0.521&0.500\\
% & Avg 
\multicolumn{2}{c|}{ETTh1}&\bf{0.422}&\bf{0.430}&0.445&0.439&\underline{0.431}&\underline{0.431}&0.450&0.440&0.458&0.450\\
\midrule

% \multirow{5}{*}{\rotatebox{90}{$ETTh2$}}
% & 96  &\underline{0.296}&\underline{0.346}&\bf{0.294}&\bf{0.343}&0.340&0.374&0.333&0.387&0.296&0.344&0.358&0.397&0.346&0.388&3.755&1.525\\
% & 192 &\bf{0.369}&\bf{0.392}&\underline{0.377}&\underline{0.393}&0.402&0.414&0.477&0.476&0.389&0.402&0.429&0.439&0.456&0.452&5.602&1.931\\
% & 336 &\bf{0.376}&\bf{0.405}&\underline{0.381}&\underline{0.409}&0.452&0.452&0.594&0.541&0.418&0.430&0.496&0.487&0.482&0.486&4.721&1.835\\
% & 720 &\bf{0.405}&\bf{0.432}&\underline{0.412}&\underline{0.471}&0.462&0.468&0.831&0.657&0.433&0.448&0.463&0.474&0.515&0.511&3.647&1.625\\
% & Avg &\bf{0.361}&\bf{0.393}&\underline{0.366}&\underline{0.404}&0.414&0.427&0.559&0.515&0.384&0.406&0.437&0.449&0.450&0.459&4.431&1.729\\
% \midrule

% \multirow{5}{*}{\rotatebox{90}{$ETTm1$}}
% & 96  &\underline{0.317}&\underline{0.356}&\bf{0.316}&\underline{0.356}&0.318&\bf{0.355}&0.321&0.360&0.338&0.375\\
% & 192 &\underline{0.357}&0.381&0.358&\underline{0.380}&\bf{0.354}&\bf{0.377}&0.362&0.384&0.371&0.387\\
% & 336 &\bf{0.387}&0.404&\underline{0.389}&\underline{0.403}&0.391&\underline{0.403}&0.392&\bf{0.402}&0.410&0.411\\
% & 720 &\bf{0.448}&0.439&\underline{0.450}&\bf{0.435}&0.453&0.437&\underline{0.450}&\bf{0.435}&0.478&0.450\\
% & Avg 
\multicolumn{2}{c|}{ETTm1}&\bf{0.377}&0.395&\underline{0.378}&\bf{0.393}&0.379&\bf{0.393}&0.381&0.395&0.400&0.406\\
\midrule
% \multicolumn{2}{c|}{Average}&\\

\bottomrule
\end{tabular}
}
\end{center}
\vskip -0.15in
\end{table*}

\begin{table}[h]
\vskip -0.10in
\caption{Utilize AttnEmbed as a plug-in. The lookback length is 336 for PatchTST and 96 for CARD. All models are averaged from 4 different horizons. A lower MSE indicates better performance. Detailed results are provided in Appendix \ref{app:plugin_full}.}
\label{tab:plugin}
\begin{center}
\scalebox{0.72}{
\begin{tabular}{c|c|cc|cc|cc|cc}
\toprule
\midrule

\multicolumn{2}{c|}{\multirow{2}{*}{Methods}}&\multicolumn{2}{c|}{\multirow{2}{*}{PatchTST(42)}}&\multicolumn{2}{c|}{PatchTST(42)}&\multicolumn{2}{c|}{\multirow{2}{*}{CARD}}&\multicolumn{2}{c}{CARD} \\
\multicolumn{2}{c|}{}&\multicolumn{2}{c|}{}&\multicolumn{2}{c|}{+AttnEmbed}&\multicolumn{2}{c|}{}&\multicolumn{2}{c}{+AttnEmbed} \\

\midrule

\multicolumn{2}{c|}{Metric} & MSE& MAE & MSE & MAE& MSE & MAE& MSE  & MAE\\
\midrule

% \multirow{5}{*}{\rotatebox{90}{$Weather$}}
% & 96  &0.152&0.199&\textbf{0.151}&\textbf{0.199}&\textbf{0.150}&\textbf{0.188}&0.152&0.190\\
% & 192 &0.197&0.243&\textbf{0.195}&\textbf{0.240}&0.202&0.238&\textbf{0.200}&\textbf{0.237}\\
% & 336 &0.249&0.283&\textbf{0.246}&\textbf{0.282}&0.260&0.282&\textbf{0.259}&\textbf{0.282}\\
% & 720 &0.320&0.335&\textbf{0.320}&\textbf{0.331}&0.343&0.353&\textbf{0.341}&\textbf{0.334}\\
% & Avg
\multicolumn{2}{c|}{Weather}&0.229&0.265&\textbf{0.227}&\textbf{0.263}&0.239&0.265&\textbf{0.238}&\textbf{0.260}\\
\midrule

% \multirow{5}{*}{\rotatebox{90}{$ETTh1$}}
% & 96  &0.375&0.399&\textbf{0.374}&\textbf{0.397}&0.383&0.391&\textbf{0.379}&\textbf{0.390}\\
% & 192 &0.414&0.421&\textbf{0.412}&\textbf{0.423}&0.435&0.420&\textbf{0.428}&\textbf{0.421}\\
% & 336 &0.431&0.436&\textbf{0.420}&\textbf{0.432}&0.479&0.442&\textbf{0.472}&\textbf{0.440}\\
% & 720 &0.449&0.466&\textbf{0.430}&\textbf{0.455}&0.471&0.461&\textbf{0.469}&\textbf{0.459}\\
% & Avg 
\multicolumn{2}{c|}{ETTh1}&0.417&0.430&\textbf{0.409}&\textbf{0.426}&0.442&0.428&\textbf{0.436}&\textbf{0.427}\\
\midrule

% \multirow{5}{*}{\rotatebox{90}{$ETTm1$}}
% & 96  &0.290&0.342&\textbf{0.286}&\textbf{0.341}&0.319&0.347&\textbf{0.316}&\textbf{0.344}\\
% & 192 &0.332&0.369&\textbf{0.331}&\textbf{0.369}&0.363&0.370&\textbf{0.356}&\textbf{0.365}\\
% & 336 &0.366&0.392&\textbf{0.363}&\textbf{0.390}&0.392&0.390&\textbf{0.386}&\textbf{0.386}\\
% & 720 &0.420&0.424&\textbf{0.410}&\textbf{0.416}&0.458&0.425&\textbf{0.450}&\textbf{0.422}\\
% & Avg 
\multicolumn{2}{c|}{ETTh2}&0.352&0.381&\textbf{0.347}&\textbf{0.379}&0.383&0.384&\textbf{0.377}&\textbf{0.379}\\
\midrule

\bottomrule
\end{tabular}}

\end{center}
\vskip -0.15in
\end{table}

\subsubsection{Utilize AttnEmbed as A Plug-in}
\label{plugin}
As depicted in Figure \ref{fig:attn_embed}, the primary distinction between AttnEmbed and PatchTST lies in the embedding module. Consequently, AttnEmbed could potentially serve as a plug-in module to replace patching.
Here, we primarily investigate two aspects of AttnEmbed's versatility: the extension of the lookback window and the incorporation of multi-channel relationships. We have smoothly incorporated the AttnEmbed module into the PatchTST framework (42) with a lookback window of 336 time steps, and into the CARD model, which utilizes a 96-step lookback window and is tailored to improve cross-channel interactions. This strategic integration underscores the adaptability and strength of our design as a robust plug-in module, demonstrating its efficacy over diverse input horizons. To ensure a fair comparison, the window size, stride settings, and lookback window are kept in line with those used in the respective original models. The results are summarized in Table~\ref{tab:plugin}. After substituting the patching with AttnEmbed, we can observe a performance improvement even when retaining the same window size and stride. Notably, on the ETTh1 dataset, AttnEmbed achieves a relative MSE reduction of $1.9\%$ when integrated with PatchTST(42), and a relative MSE reduction of $1.3\%$ for CARD. This indicates that AttnEmbed is versatile beyond the constraints of lookback window and channel-independent settings, demonstrating its potential as an adaptable plug-in module for various models.

\subsection{Model Analysis}

\subsubsection{Ablations}

Here, we carry out ablation studies for the architectural design of AttnEmbed, with the aim of demonstrating the performance impact of omitting the global landmark or EMA components. Two ablated versions of AttnEmbed are evaluated on the ETTh1 and ETTm1 datasets: 1) AttnEmbed without global landmark, to assess the significance of incorporating global information; and 2) AttnEmbed without EMA, to ascertain the contribution of EMA to time series forecasting. 
As depicted in Table \ref{tab:ablation_ema_landmark}, the fully-equipped AttnEmbed model, which integrates both landmark and EMA, outperforms its two ablated variants by achieving an average MSE reduction of $2.2\%$. This highlights the crucial roles that landmarks and EMA play within the AttnEmbed framework, effectively capturing global information and local time-dependent dynamics.

\begin{table}[h]
\vskip -0.10in
\caption{Ablation on EMA and landmark with a lookback length of 96. All models are averaged from 4 different horizons. A lower MSE indicates better performance. Detailed results are provided in Appendix \ref{app:ablation_full}}
\label{tab:ablation_ema_landmark}
\begin{center}
\scalebox{0.8}{

\begin{tabular}{c|c|cc|cc|cc}
\toprule
\midrule

\multicolumn{2}{c|}{\multirow{2}{*}{Methods}}&\multicolumn{2}{c|}{\multirow{2}{*}{AttnEmbed}}&\multicolumn{2}{c|}{AttnEmbed}&\multicolumn{2}{c}{AttnEmbed} \\
\multicolumn{2}{c|}{}&\multicolumn{2}{c|}{}&\multicolumn{2}{c|}{w/o EMA }&\multicolumn{2}{c}{w/o Landmark} \\

\midrule

\multicolumn{2}{c|}{Metric} & MSE& MAE & MSE & MAE& MSE & MAE\\
\midrule

% \multirow{5}{*}{\rotatebox{90}{$ETTh1$}}
% & 96  &0.367&0.398&0.378&0.398&0.370&0.400\\
% & 192 &0.420&0.428&0.423&0.425&0.421&0.427\\
% & 336 &0.448&0.438&0.456&0.430&0.440&0.436\\
% & 720 &0.454&0.459&0.472&0.455&0.461&0.454\\
% & Avg 
\multicolumn{2}{c|}{ETTh1}&0.422&0.430&0.432&0.427&0.423&0.429\\
\midrule

% \multirow{5}{*}{\rotatebox{90}{$ETTm1$}}
% & 96  &0.317&0.356&0.319&0.360&0.326&0.365\\
% & 192 &0.357&0.381&0.374&0.391&0.370&0.384\\
% & 336 &0.387&0.404&0.396&0.402&0.396&0.403\\
% & 720 &0.448&0.439&0.449&0.437&0.455&0.436\\
% & Avg 
\multicolumn{2}{c|}{ETTm1}&0.377&0.395&0.384&0.397&0.386&0.397\\
\midrule

\bottomrule
\end{tabular}
}
\end{center}
\vskip -0.15in
\end{table}

\subsubsection{Parameter Analysis}
\begin{figure}[h]
    \centering
    \includegraphics[width=0.48\textwidth]{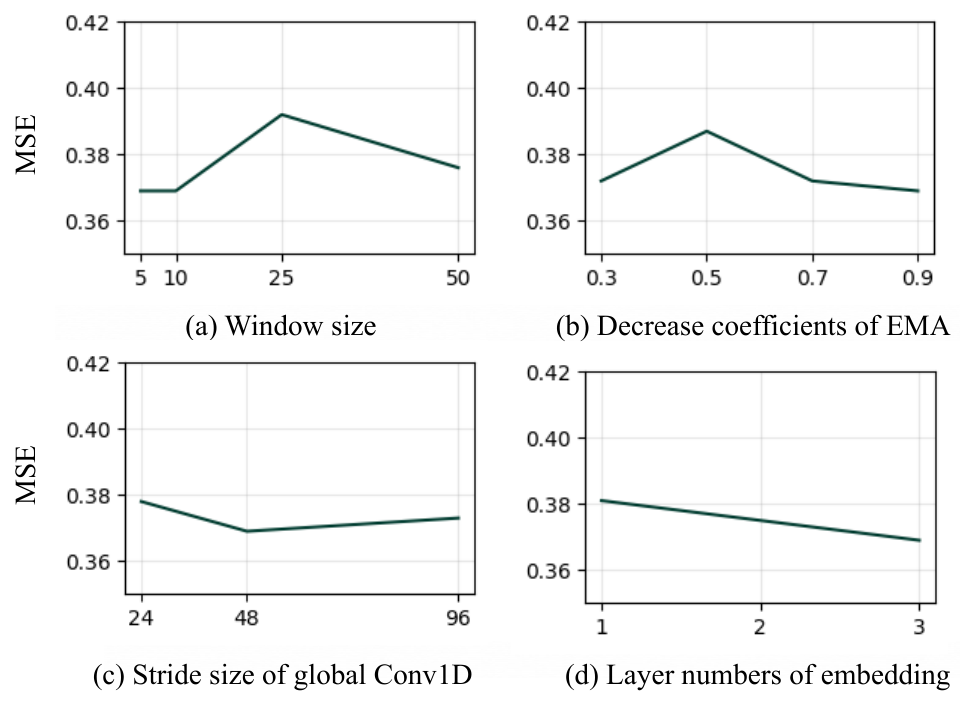}
    \vskip -0.1in
    \caption{Parameter analysis on ETTh1 with a lookback window of 96 and a horizon of 96.
    % of window size, decrease coefficients of EMA, stride sizes of global Conv1D and layer numbers of the attention embedding module.
    (a) Window size. (b) Decrease coefficients of EMA. (c) Stride sizes of global Conv1D. (d) Layer numbers of the attention embedding module.
    % (a) Window size ([5, 10, 25, 50]). (b) Decrease coefficients of EMA ([0.3, 0.5, 0.7, 0.9]). (c) Stride sizes of global Conv1D ([24, 48, 96]). (d) Layer numbers of the attention embedding module ([1, 2, 3]).
    }
    \label{fig:param_analysis}
    \vskip -0.25in
\end{figure}

We examined AttnEmbed's parameter sensitivity, presenting the forecasting MSE for varying configurations in Figure \ref{fig:param_analysis}. These parameters include window sizes ([5, 10, 25, 50]), EMA decay coefficients ([0.3, 0.5, 0.7, 0.9]), Conv1D stride sizes ([24, 48, 96]), and attention embedding module layers ([1, 2, 3]), all tested on the ETTh1 dataset with a 96-period lookback and forecast horizon.

Figure \ref{fig:param_analysis}(a) reveals that larger window sizes struggle with quick distribution changes. In contrast, as shown in Figure \ref{fig:param_analysis}(b), the proper EMA decay coefficient can enhance results and mitigate noise, although too low a coefficient may over-smooth and degrade performance. Figure \ref{fig:param_analysis}(c) suggests that a stride size around half the lookback window optimizes the capture of global patterns. Lastly, Figure \ref{fig:param_analysis}(d) indicates that deeper attention embedding layers improve outcomes, with three layers being selected for their balance of performance and computational efficiency.

% We also investigate the parameter sensitivity of AttnEmbed. Specifically, we report the forecasting MSE under different parameter settings in  Figure \ref{fig:param_analysis}, including window sizes ([5, 10, 25, 50]), decrease coefficients of EMA ([0.3, 0.5, 0.7, 0.9]), stride sizes of global Conv1D ([24, 48, 96]), and layer numbers of the attention embedding module ([1, 2, 3]). The aforementioned parameters are evaluated using the ETTh1 dataset, with a lookback window of 96 and a horizon of 96. 
% As depicted in Figure \ref{fig:param_analysis} (a), larger window sizes perform poorly in capturing rapid distribution shifts. 
% Figure \ref{fig:param_analysis} (b) indicates that the suitable application of EMA can effectively refine results and reduce noise. However, an excessively low decay coefficient can lead to over-smoothing and consequently impact performance.
% Figure \ref{fig:param_analysis} (c) illustrates that setting the stride size to approximately half of the lookback window can more effectively capture global information.
% Figure \ref{fig:param_analysis} (d) demonstrates that the deep layer of the attention embedding module can enhance performance. In order to optimize time efficiency, we choose to use a parameter value of 3.

\subsubsection{Alleviating Rank Collapse}

Rank collapse is a notable challenge in the application of transformer models, wherein the attention matrix's rank decreases during training. This contraction in rank can constrain the model's ability to fit data, leading to weaker generalization. Although this issue affects transformers for time series (TS), they are less prone to it compared to the more layered Large Language Models (LLMs), yet it remains a relevant concern for TS model robustness.Since attention matrix is closely related to kernel matrix that often exhibits a higher rank than the input matrix, we expect the introduction of attention based representation may help alleviate the problem of rank collapse. 

Though following \cite{Dong2021AttentionIN}, we compare the relative norm of the residual to evaluate the `rankness' of layers in a model. As shown in Figure~\ref{fig:attn_embed}, using the attention based representation,  where pairwise similarities are computed using RBF and polynomial kernels, can efficiently mitigate the issue of rank collapse observed in time series data. More information about kernel function for attention based representation can be found in Section \ref{sec:kernel_function}. 

\begin{figure}[h]
    \centering
    \includegraphics[width=0.48\textwidth]{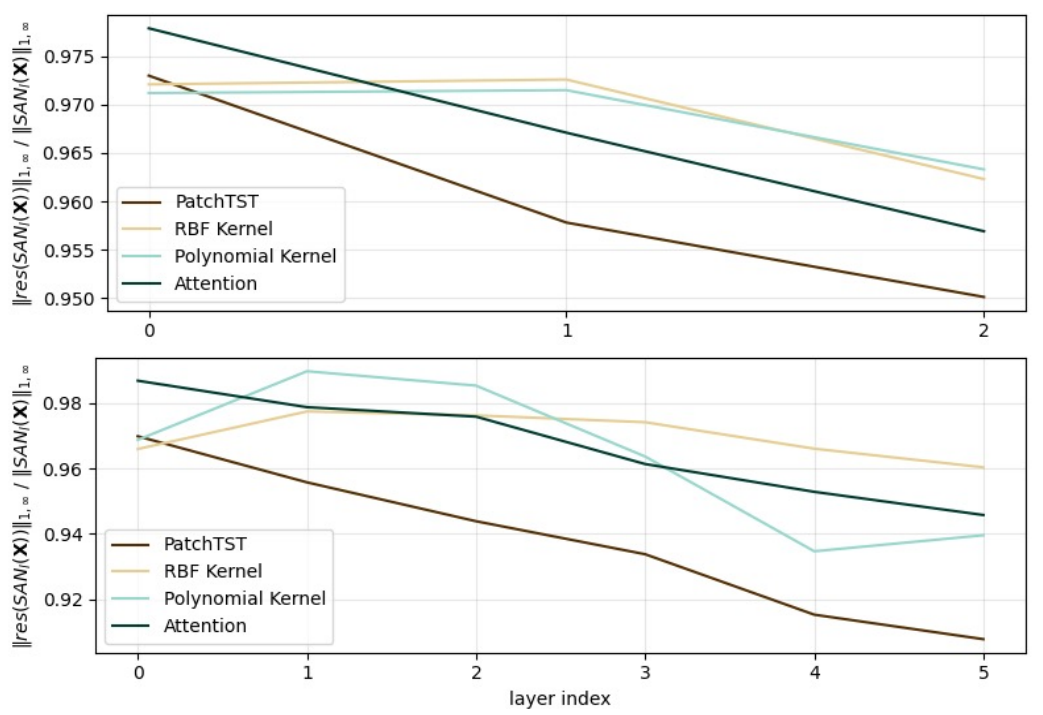}
    \vskip -0.15in
    \caption{\small Relative norm of the residual along the depth for PatchTST, Attention, RBF kernel and polynomial kernel with different layers ([3, 6]) of transformer encoder on ETTh1. 
    % The experimental setup, including the lookback length, horizon length, window (or patch) size, and stride size, is kept consistent. The rank of PatchTST decays more precipitously than other methods.
    }
    \label{fig:attn_embed}
    \vskip -0.15in
\end{figure}

\section{Conclusion}
\label{sec_conclusion}
The paper addresses the inherent nature of time series data, such as its low information density and the prevalence of distribution shifts and noises. By leveraging an attention mechanism tailored to time series, where the attention weights play a central role in representing data, we propose a novel and robust embedding strategy that utilizes global landmarks and a localized window to enrich the data representation. Our tailored attention map significantly outperforms the patching embedding-based SOTA transformer model in time series forecasting, a testament to its effectiveness. The results are compelling, with our approach yielding an average 3.6\% improvement in MSE for SOTA multivariate time series prediction. The enhancement is designed to elevate predictive precision and introduces a modular component that is engineered for seamless integration within existing architectures, potentially reinforcing their resilience in generating embeddings from noisy signals.

\newpage
\section*{Impact Statements}
This paper presents work whose goal is to advance the field of Machine Learning. There are many potential societal consequences of our work, none which we feel must be specifically highlighted here.
\bibliography{sample-base}
\bibliographystyle{icml2024}

%%%%%%%%%%%%%%%%%%%%%%%%%%%%%%%%%%%%%%%%%%%%%%%%%%%%%%%%%%%%%%%%%%%%%%%%%%%%%%%
%%%%%%%%%%%%%%%%%%%%%%%%%%%%%%%%%%%%%%%%%%%%%%%%%%%%%%%%%%%%%%%%%%%%%%%%%%%%%%%
% APPENDIX
%%%%%%%%%%%%%%%%%%%%%%%%%%%%%%%%%%%%%%%%%%%%%%%%%%%%%%%%%%%%%%%%%%%%%%%%%%%%%%%
%%%%%%%%%%%%%%%%%%%%%%%%%%%%%%%%%%%%%%%%%%%%%%%%%%%%%%%%%%%%%%%%%%%%%%%%%%%%%%%
\newpage
\onecolumn
\newpage
\appendix
\section{Full theoretical Analysis of attention as Robust Representation}
\label{app:theoretical}
In this section, we will show that using attention map as an alternative representation can be significantly more robust than the original inputs, particularly to the noise. In other words, attention map help reduce the impact of noises compared to the original inputs. 

Let $x_i \in \R^d, i =1, \ldots, n$ be $n$ vectors. For simplicity of analysis, we assume that each vector is generated from one of $m < d$ Gaussian distributions, denoted by $\N\left(\mu_i, I_d\right), i=1, \ldots, m$. For the convenience of analysis, we assume that $\langle \mu_i, \mu_j \rangle = \delta_{i,j}s$ for any $i,j \in [m]$. Let $n = m K$, and we choose to generate $K$ vectors from each Gaussian distributions. In particular, vector $x_{mj+i}$, with $j=0,\ldots, K-1, i=1,\ldots, m$, is generated from Gaussian distribution $\N\left(\mu_{i}, I_d\right)$. Then, if we choose two vectors $x_i^+$ and $x_j^+$ that are sampled from the same Gaussian distribution, we have
\[
\E[|x^+_i - x^+_j|^2] = 2d
\]
and if $x^-_i$ and $x^-_j$ are sampled from different distributions, we have
\[
\E[|x^-_i - x^-_j|^2] = 2d + 2s
\]
It is clearly that the relative difference between the two expected distance square is $O(s/d)$. In fact, we can further show that there is a significant chance for the distance between two data points sampled from the same distributions to be larger than the distance between two data points sampled from different distributions, implying that the added noises can significantly affect the geometrical relationship among the sampled data points. To this end, we write $x_i^+ = \mu_a + z^+_i$ and $x_j^+ = \mu_a + z^+_j$, where $z^+_i, z^+_j \sim \N(0, I_d)$. Hence
\[
|x_i^+ - x_j^+|^2 = |\underbrace{z^+_i - z^+_j}_{:=u_+}|^2
\]
It is clear that $|u_+|^2 \sim 2\chi^2_d$. In the meantime, by writing $x_i^- = \mu_b + z_i^-$ and $x_j^- = \mu_c + z_j^-$, with $z_i^-, z_j^- \sim \N(0, I_d)$, we have
\[
|x_i^- - x_j^-|^2 = 2s + 2\underbrace{\langle \mu_b - \mu_c, z_i^{-} - z_j^-\rangle}_{:= v_-} + |\underbrace{z_i^- - z_j^-}_{:= u_-}|^2
\]
It is clear that $v_-\sim \N(0, 2s)$ and $|u_-|^2\sim2\chi_d^2$. We want to bound the probability 
\[
\Pr\left(|u_-|^2 + 2v_- - |u_+|^2 \leq -2s\right)
\]
First, using the standard concentration inequality for $\chi^2_d$ distributions, we have
\[
\Pr\left(\frac{1}{2}|u_-|^2 \leq d + 2\sqrt{d\delta} + 2\delta\right) \geq 1 - \exp(-\delta)
\]
By setting $\delta = s^2/(16d)$, under the assumption $s^2 \leq 16d$, we have
\[
\Pr\left(|u_-|^2 \leq 2d + s + \frac{s^2}{4d}\right) \geq 1 - \exp\left(-\frac{s^2}{16 d}\right) \geq \frac{s^2}{32 d}
\]
Since $v_-\sim\N(0, 2s)$, we have
\[
\Pr\left(v_- \geq \frac{2s}{3}\right) \leq \frac{3}{2\sqrt{2\pi}s}\exp\left(-\frac{s}{9}\right) \geq \frac{e^{-s/9}}{\sqrt{2}s}
\]
and therefore
\[
\Pr\left(|u_+|^2 + 2v_- \leq 2d + \frac{7s}{3} + \frac{s^2}{4d}\right) \leq \frac{s^2}{32d} - \frac{e^{-s/9}}{\sqrt{2}s}
\]
In the meantime, we can also lower bound $|u_+|^2$. Using the fact the CDF for $\chi^2_2$ is $1 - \exp(-x/2)$. We have
\[
\Pr\left(|u_+|^2 \geq 2(1 + \varepsilon) d\right) \geq \exp\left(-\frac{\varepsilon d}{2}\right)
\]
By choosing $\varepsilon = 3s/(2d)$, we have
\[
\Pr\left(|u_+|^2 \geq 2d + 3s \right) \geq \exp\left(-\frac{3s}{4}\right)
\]
Combining the above two inequalities, we have
\[
\Pr\left(|u_-|^2 - |u_+|^2 + 2v_- \leq -\frac{2s}{3} + \frac{s^2}{4d}\right)\geq \exp\left(-\frac{3s}{4}\right)\left(\frac{s^2}{32d} - \frac{e^{-s/9}}{\sqrt{2}s}\right)
\]
when
\[
s < d \leq \frac{\sqrt{2}s^3}{32}
\]
we have
\[
\Pr\left(|u_-|^2 - |u_+|^2 + 2v_- \leq -\frac{s}{3}\right)\geq \frac{1}{\sqrt{2}s}\exp\left(-\frac{3s}{4}\right)\left(1 - e^{-s/9}\right)
\]
implying that there is a descent chance for $|x_i^- - x_j^-| < |x_i^+ - x_j^+|$. 

Now, let's check the attention based representation, i.e. for any vector $x$, we represent it by $f(x)$ given below
\[
f(x) = \left(\exp\left(\lambda\langle x, x_1\rangle\right), \ldots, \exp\left(\lambda \langle x, x_n\rangle\right) \right)
\]
For simplicity, we sample a pair of data points $x_i^+, x_j^+\sim \N(\mu_a, I_d)$ from the same distribution, and another pair of data points $x_i^-\sim\N(\mu_b, I_d)$ and $x_j^- \sim \N(\mu_c, I_d)$. We first represent each of these four data points by their attention map. We first compute the distance
\begin{eqnarray*}
\E[|f(x_i^+) - f(x_j^+)|^2]
& = & \sum_{k=1}^n \E\left[\left|\exp\left(\lambda\langle x_i^+, x_k\rangle\right) - \exp\left(\lambda \langle x_j^+, x_k\rangle\right)\right|^2\right] \\
& = & K\sum_{k=1}^m \E\left[\left|\exp\left(\lambda\langle x_i^+, x\rangle\right) - \exp\left(\lambda \langle x_j^+, x\rangle\right)\right|^2\right] \\
& = & K\sum_{k=1}^m\E\left[2\exp\left(2\lambda\langle x_i^+, x\rangle\right) - 2\exp\left(\lambda\langle x, x_i^+ + x_j^+\rangle \right)\right] 
\end{eqnarray*}
To compute the above expectation, we first consider $k=a$. Define $z_i^+ = x_i^+ - \mu_a$, $z_j^+ = x_j^+ - \mu_a$, and $z = x - \mu_a$. We have
\begin{eqnarray*}
\lefteqn{\E\left[2\exp\left(2\lambda\langle x_i^+, x\rangle\right) - 2\exp\left(\sqrt{2}\langle x, x_j^+\rangle\right)\right]} \\
& = & \E\left[2\exp\left(2\lambda(s+\langle z, z_i^+\rangle + \langle \mu_a, z_i^+ + z\rangle )\right) - 2\exp\left(\sqrt{2}\lambda\left(\sqrt{2}s+\langle z, z_j^+\rangle + \langle \mu_a, z_j^+ + z\rangle \right)\right)\right]
\end{eqnarray*}
By taking the expectation over $z_i^+$ and $z_j^+$, given $\langle z_i^+, z + \mu_a\rangle \sim \N(0, |z+\mu_a|^2)$ and $z_j^+$ and $\langle z_j^+, z + \mu_a\rangle \sim \N(0, |z+\mu_a|^2)$, we have
\[
\E_{z_i^+}\left[\exp\left(2\lambda\langle z_i^+, z + \mu_a\rangle\right)\right] = \exp\left(2\lambda^2|z+\mu_a|^2\right), \; \E_{z_j^+}\left[\exp\left(\sqrt{2}\lambda\langle z_j^+, z + \mu_a\rangle\right)\right] = \exp\left(\lambda^2|z+\mu_a|^2\right)
\]
and therefore
\begin{eqnarray*}
\lefteqn{\E\left[2\exp\left(2\lambda\langle x_i^+, x\rangle\right) - 2\exp\left(\sqrt{2}\langle x, x_j^-\rangle\right)\right]} \\
& = & \E\left[\exp\left(2\lambda\left(s+\langle \mu_a, z\rangle + \lambda|z+\mu_a|^2\right)\right)\right] - \E\left[\exp\left(\sqrt{2}\lambda\left(s+\langle \mu_a, z\rangle + \frac{\lambda}{\sqrt{2}}|z+\mu_a|^2\right)\right)\right]
\end{eqnarray*}
Since
\begin{eqnarray*}
\lefteqn{\E\left[\exp\left(2\lambda^2|z+\mu_a|^2 + 2\lambda \langle \mu_a, z\rangle \right)\right]} \\
& = & \frac{e^{2\lambda^2s}}{(2\pi)^{d/2}}\int \exp\left(2\lambda^2|z|^2 + 2\lambda(2\lambda + 1)\langle\mu_a, z\rangle - \frac{|z|^2}{2}\right) dz \\
& = & \frac{e^{2\lambda^2s}}{(2\pi)^{(d-1)/2}}\int \exp\left(-(1 - 4\lambda^2)\frac{|z_{d-1}|^2}{2}\right) dz_{d-1} \times \frac{1}{\sqrt{2\pi}}\int\exp\left(-(1-4\lambda^2)\frac{z^2}{2} + 2\lambda(2\lambda + 1) s^{1/2} z\right) \\
& = & \frac{e^{2\lambda^2s}}{(1 - 4\lambda^{2})^{d/2}}\exp\left(\frac{2\lambda^2(1+2\lambda)^2s}{1 - 4\lambda^2}\right) = \frac{e^{2\lambda^2s}}{(1 - 4\lambda^{2})^{d/2}}\exp\left(\frac{2\lambda^2(1+2\lambda)s}{1 - 2\lambda}\right) := C_1
\end{eqnarray*}
and
\begin{eqnarray*}
\lefteqn{\E\left[\exp\left(\sqrt{2}\lambda\langle \mu_a, z\rangle + \lambda^2|z + \mu_a|^2\right)\right]} \\
& = & \frac{e^{\lambda^2s}}{(2\pi)^{d/2}}\int\exp\left(-(1 - 2\lambda^2)\frac{|z|^2}{2} + \lambda(\sqrt{2} + 2\lambda)\langle z, \mu_a\rangle\right) dz \\
& = & \frac{e^{\lambda^2 s}}{(1-2\lambda^2)^{d/2}}\exp\left(\frac{\lambda^2(1+\sqrt{2}\lambda)^2s}{2(1 - 2\lambda^2)}\right) = \frac{e^{\lambda^2 s}}{(1-2\lambda^2)^{d/2}}\exp\left(\frac{\lambda^2(1+\sqrt{2}\lambda)s}{2(1 - \sqrt{2}\lambda)}\right) := C_2
\end{eqnarray*}
we have
\begin{eqnarray*}
\lefteqn{\E\left[2\exp\left(2\lambda\langle x_i^+, x\rangle\right) - 2\exp\left(\langle x, x_i^++x_j^+\rangle\right)\right]} \\
& = & \frac{2e^{2\lambda(1+\lambda)s}}{(1 - 4\lambda^{2})^{d/2}}\exp\left(\frac{2\lambda^2(1+2\lambda)s}{1 - 2\lambda}\right) - \frac{2e^{\lambda(2 + \lambda) s}}{(1-2\lambda^2)^{d/2}}\exp\left(\frac{\lambda^2(1+\sqrt{2}\lambda)s}{2(1 - \sqrt{2}\lambda)}\right)
\end{eqnarray*}
For $k \neq a$, we have
\begin{eqnarray*}
\lefteqn{\E\left[2\lambda\exp\left( 2\lambda\langle x_i^+, x\rangle \right) - 2\exp\left(\sqrt{2}\langle x, x_j^+\rangle\right) \right]} \\
& = & \frac{2e^{2\lambda^2s}}{(1 - 4\lambda^{2})^{d/2}}\exp\left(\frac{2\lambda^2(1+2\lambda)s}{1 - 2\lambda}\right) - \frac{2e^{\lambda^2 s}}{(1-2\lambda^2)^{d/2}}\exp\left(\frac{\lambda^2(1+\sqrt{2}\lambda)s}{2(1 - \sqrt{2}\lambda)}\right)
\end{eqnarray*}
By combining the above results, we have
\begin{eqnarray*}
\E\left[\left|f(x_i^+) - f(x_j^+)\right|^2\right] = 2K\left(e^{2\lambda s} + m - 1\right) \left(C_1 - C_2\right)
\end{eqnarray*}

We then compute the distance between $f(x_i^-)$ and $f(x_j^-)$, which is given by
\begin{eqnarray*}
\E\left[|f(x_i^-) - f(x_j^-)|^2\right] = K\sum_{k=1}^m\E\left[\exp\left(2\lambda\langle x_i^-, x\rangle\right) + \exp\left(2\lambda\langle x_j^-, x\rangle\right) - 2\exp\left(\lambda\langle x, x_i^- + x_j^-\rangle \right)\right] 
\end{eqnarray*}
First, for the case $k = b$, we have
\begin{eqnarray*}
\lefteqn{\E\left[\exp\left(2\lambda\langle x_i^+, x\rangle\right) + \exp\left(2\lambda\langle x_j^+, x\rangle\right) - 2\exp\left(\lambda\langle x, x_i^+ + x_j^+\rangle \right)\right] } \\
& = & \E\left[\exp\left(2\lambda s + 2\lambda \langle \mu_b, z+z_i^-\rangle + 2\lambda\langle z, z_i^-\rangle\right) + \exp\left(2\langle \mu_b, z\rangle + 2\lambda \langle \mu_c, z_i^-\rangle + 2\lambda\langle z, z_i^-\rangle\right)\right] \\
& & 
- 2\E\left[\exp\left(\lambda s + \lambda\langle \mu_b, z_i^{-} + z_j^-\rangle + \lambda\langle \mu_b + \mu_c, z_j\rangle + \lambda \langle z, z_i^- + z_j^-\rangle\right)\right] \\
& = & \E\left[\exp\left(2\lambda s + 2\lambda^2|z + \mu_b|^2 + 2\lambda\langle \mu_b, z\rangle\right)\right] + \E\left[\exp\left(2\lambda\langle\mu_b, z\rangle + 2\lambda^2|\mu_c + z|^2\right)\right] \\
& & - 2\E\left[\exp\left( \lambda s + \lambda \langle \mu_b + \mu_c, z_j^-\rangle + \frac{\lambda^2|z_i^- + z_j^-|^2}{2}\right)\right] \\
& = & e^{2\lambda s}C_1 + \E\left[\exp\left(2\lambda^2s + 2\lambda\langle\mu_b + 2\lambda\mu_c, z\rangle + 2\lambda^2 |z|^2\right)\right] - 2\E\left[\exp\left(\lambda s+ \lambda\langle \mu_b + \mu_c, z_j^-\rangle + \lambda^2|z_i^-|^2\right)\right] \\
& = & e^{2\lambda s}C_1 + \frac{e^{2\lambda^2s}}{(1 - 4\lambda^2)^{d/2}}\exp\left(\frac{2\lambda^2(1+4\lambda^2)s}{1 - 4\lambda^2}\right) - \frac{2e^{\lambda s+ \lambda^2 s}}{(1 - 2\lambda^2)^{d/2}} \\
& \geq & \left(e^{2\lambda s} + e^{-8\lambda^3}\right) C_1 - 2\exp\left(-\frac{\lambda^2s}{2}\right)C_2
\end{eqnarray*}
Second, for $k \neq b$ and $k \neq c$, we have
\begin{eqnarray*}
\lefteqn{\E\left[\exp\left(2\lambda\langle x_i^+, x\rangle\right) + \exp\left(2\lambda\langle x_j^+, x\rangle\right) - 2\exp\left(\lambda\langle x, x_i^+ + x_j^+\rangle \right)\right] } \\
& = & 2\E\left[ \exp\left(2\langle \mu_b, z\rangle + 2\lambda \langle \mu_c, z_i^-\rangle + 2\lambda\langle z, z_i^-\rangle\right)\right] \\
& & 
- 2\E\left[\exp\left(\lambda\langle \mu_k, z_i^{-} + z_j^-\rangle + \lambda\langle \mu_b + \mu_c, z_j\rangle + \lambda \langle z, z_i^- + z_j^-\rangle\right)\right] \\
& = & 2\E\left[\exp\left(2\lambda\langle\mu_b, z\rangle + 2\lambda^2|\mu_c + z|^2\right)\right] - 2\E\left[\exp\left(\sqrt{2}\lambda\langle \mu_k + z, z_i^-\rangle + \lambda \langle \mu_b + \mu_c, z_j^- \rangle\right)\right] \\
& = &  2\E\left[\exp\left(2\lambda^2s + 2\lambda\langle\mu_b + 2\lambda\mu_c, z\rangle + 2\lambda^2 |z|^2\right)\right] - 2\E\left[\exp\left(\lambda^2 s + \lambda^2|\mu_k + z|^2\right)\right] \\
& = & \frac{2e^{2\lambda^2s}}{(1 - 4\lambda^2)^{d/2}}\exp\left(\frac{2\lambda^2(1+4\lambda^2)}{1 - 4\lambda^2}\right) - \frac{2e^{\lambda^2 s}}{(1 - 2\lambda^2)^{d/2}} \\
& \geq & 2e^{-8\lambda^3} C_1 - 2\exp\left(-\frac{\lambda^2s}{2}\right)C_2
\end{eqnarray*}
Thus, we have
\[
\E\left[|f(x_i^-) - f(x_j^-)|^2\right] \geq K\left(2\left(e^{2\lambda s} + m e^{-8\lambda^3}\right)C_1 - m \exp\left(-\frac{\lambda^2s}{2}\right)C_2\right) \geq 2Ke^{2\lambda s} C_1
\]
and hence
\[
\frac{\E\left[|f(x_i^-) - f(x_i^-)|^2\right] - \E\left[|f(x_i^+) - f(x_i^+)|^2\right]}{\E\left[|f(x_i^+) - f(x_i^+)|^2\right]} \geq \frac{C_1/(C_1 - C_2) - \left(1 + (m - 1)e^{-2\lambda s}\right)}{1 + (m - 1)e^{-2\lambda s}}
\]
Since
\[
\frac{C_1}{C_1 - C_2} = \frac{1}{1 - \underbrace{e^{-\lambda^2s}\left(\frac{1 - 4\lambda^2}{1 - 2\lambda^2}\right)^{d/2}\exp\left(\lambda^2s\left[ \frac{1 + \sqrt{2}\lambda}{2(1 - \sqrt{2}\lambda)}-\frac{2(1+2\lambda)}{1-2\lambda}\right]\right)}_{:=\Gamma}}
\]
It is easy to verify that when $\lambda^2 = 1/d$, $\Gamma = \Omega(1)$, and by further assuming that $s$ is sufficiently large that $e^{-2\lambda s} \leq \gamma/(2(m-1))$, we have
\[
\frac{\E\left[|f(x_i^-) - f(x_i^-)|^2\right] - \E\left[|f(x_i^+) - f(x_i^+)|^2\right]}{\E\left[|f(x_i^+) - f(x_i^+)|^2\right]} \geq \frac{\Gamma}{2 + \Gamma} \geq \frac{\Gamma}{3}
\]
This analysis shows that by using attention as a representation, we are able to easily distinguish if data points come from the same distribution, even with a very large noise, which become difficult if we use the original inputs. 

\section{Detailed Results}

\subsection{Detailed Results of Multivariate Forecasting}
\label{app:full_multi}
\begin{table*}[h]
    \caption{Multivariate forecasting results. The lookback length is set as 96. All models are evaluated on 4 different prediction horizons \{96, 192, 336, 720\}. A lower MSE indicates better performance. The best ones are in Bold, and the second ones are \underline{underlined}.}
    \label{tab:multi_forecasting}
    \begin{center}
    \scalebox{0.8}{
    \small
    \begin{tabular}{c|c|cc|cc|cc|cc|cc|cc|cc|cc}
    \toprule
    \midrule
    
    \multicolumn{2}{c|}{Methods}&\multicolumn{2}{c|}{\bf{AttnEmbed}}&\multicolumn{2}{c|}{PatchTST}&\multicolumn{2}{c|}{TimesNet}&\multicolumn{2}{c|}{DLinear}&\multicolumn{2}{c|}{FiLM}&\multicolumn{2}{c|}{FEDformer}&\multicolumn{2}{c|}{Autoformer}&\multicolumn{2}{c}{Informer} \\
    
    \midrule
    
    \multicolumn{2}{c|}{Metric} & MSE& MAE & MSE & MAE& MSE & MAE& MSE  & MAE& MSE  & MAE& MSE  & MAE& MSE  & MAE & MSE  & MAE \\
    \midrule
    
    \multirow{5}{*}{\rotatebox{90}{$Weather$}}
    & 96  &\bf{0.171}&\bf{0.215}&0.178&\underline{0.219}&\underline{0.172}&0.220&0.196&0.255&0.193&0.234&0.217&0.296&0.266&0.336&0.300&0.384\\
    & 192 &\bf{0.218}&\bf{0.257}&0.224&\underline{0.259}&\underline{0.219}&0.261&0.237&0.296&0.236&0.269&0.276&0.336&0.307&0.367&0.598&0.544\\
    & 336 &\bf{0.274}&\bf{0.297}&\underline{0.278}&\underline{0.298}&0.280&0.306&0.283&0.335&0.288&0.304&0.339&0.380&0.359&0.395&0.578&0.523\\
    & 720 &\underline{0.348}&\bf{0.346}&0.350&\bf{0.346}&0.365&0.359&\bf{0.345}&0.381&0.358&0.350&0.403&0.428&0.419&0.428&1.059&0.741\\
    & Avg &\bf{0.252}&\bf{0.278}&\underline{0.257}&\underline{0.280}&0.259&0.287&0.265&0.317&0.269&0.339&0.309&0.360&0.338&0.382&0.634&0.548\\
    \midrule
    
    \multirow{5}{*}{\rotatebox{90}{$ETTh1$}}
    & 96  &\bf{0.367}&\bf{0.398}&0.393&0.408&0.384&0.402&0.386&\underline{0.400}&0.388&0.401&\underline{0.376}&0.419&0.449&0.459&0.865&0.713\\
    & 192 &\bf{0.420}&\bf{0.428}&0.445&0.434&\underline{0.436}&\underline{0.429}&0.437&0.432&0.443&0.439&\bf{0.420}&0.448&0.500&0.482&1.008&0.792\\
    & 336 &\bf{0.448}&\bf{0.438}&0.484&\underline{0.451}&0.491&0.469&0.481&\underline{0.459}&0.488&0.466&0.459&0.465&0.521&0.496&1.107&0.809\\
    & 720 &\bf{0.454}&\bf{0.459}&\underline{0.480}&\underline{0.471}&0.521&0.500&0.519&0.516&0.525&0.519&0.506&0.507&0.514&0.512&1.181&0.865\\
    & Avg &\bf{0.422}&\bf{0.430}&0.450&\underline{0.440}&0.458&0.450&0.456&0.452&0.461&0.456&\underline{0.440}&0.460&0.496&0.487&1.040&0.795\\
    \midrule
    
    \multirow{5}{*}{\rotatebox{90}{$ETTh2$}}
    & 96  &\underline{0.296}&\underline{0.346}&\bf{0.294}&\bf{0.343}&0.340&0.374&0.333&0.387&0.296&0.344&0.358&0.397&0.346&0.388&3.755&1.525\\
    & 192 &\bf{0.369}&\bf{0.392}&\underline{0.377}&\underline{0.393}&0.402&0.414&0.477&0.476&0.389&0.402&0.429&0.439&0.456&0.452&5.602&1.931\\
    & 336 &\bf{0.376}&\bf{0.405}&\underline{0.381}&\underline{0.409}&0.452&0.452&0.594&0.541&0.418&0.430&0.496&0.487&0.482&0.486&4.721&1.835\\
    & 720 &\bf{0.405}&\bf{0.432}&\underline{0.412}&\underline{0.471}&0.462&0.468&0.831&0.657&0.433&0.448&0.463&0.474&0.515&0.511&3.647&1.625\\
    & Avg &\bf{0.361}&\bf{0.393}&\underline{0.366}&\underline{0.404}&0.414&0.427&0.559&0.515&0.384&0.406&0.437&0.449&0.450&0.459&4.431&1.729\\
    \midrule
    
    \multirow{5}{*}{\rotatebox{90}{$ETTm1$}}
    & 96  &\bf{0.317}&\bf{0.356}&\underline{0.321}&\underline{0.360}&0.338&0.375&0.345&0.372&0.348&0.367&0.379&0.416&0.505&0.475&0.672&0.571\\
    & 192 &\bf{0.357}&\bf{0.381}&\underline{0.362}&\underline{0.384}&0.371&0.387&0.380&0.389&0.387&0.385&0.426&0.441&0.553&0.496&0.795&0.669\\
    & 336 &\bf{0.387}&\underline{0.404}&\underline{0.392}&\bf{0.402}&0.410&0.411&0.413&0.413&0.418&0.405&0.445&0.459&0.621&0.537&1.212&0.871\\
    & 720 &\bf{0.448}&\underline{0.439}&\underline{0.450}&\bf{0.435}&0.478&0.450&0.474&0.453&0.479&0.440&0.543&0.490&0.671&0.561&1.166&0.823\\
    & Avg &\bf{0.377}&\bf{0.395}&\underline{0.381}&\bf{0.395}&0.400&0.406&0.403&0.407&0.408&0.399&0.448&0.452&0.588&0.517&0.961&0.734\\
    \midrule
    
    \multirow{5}{*}{\rotatebox{90}{$ETTm2$}}
    & 96  &\underline{0.181}&\underline{0.265}&\bf{0.178}&\bf{0.260}&0.187&0.267&0.193&0.292&0.183&0.266&0.203&0.287&0.255&0.339&0.365&0.453\\
    & 192 &\bf{0.245}&\bf{0.304}&0.249&0.307&0.249&0.309&0.284&0.362&\underline{0.247}&\underline{0.305}&0.269&0.328&0.281&0.340&0.533&0.563\\
    & 336 &\bf{0.309}&0.349&0.313&\underline{0.346}&0.321&0.351&0.369&0.427&\bf{0.309}&\bf{0.343}&0.325&0.366&0.339&0.372&1.363&0.887\\
    & 720 &0.409&0.407&\bf{0.400}&\bf{0.398}&0.408&0.403&0.554&0.522&\underline{0.407}&\bf{0.398}&0.421&0.415&0.433&0.432&3.379&1.338\\
    & Avg &\underline{0.286}&0.331&\bf{0.285}&\bf{0.327}&0.291&0.333&0.350&0.401&0.287&\underline{0.328}&0.305&0.349&0.327&0.371&1.410&0.810\\
    \midrule
    
    \multirow{5}{*}{\rotatebox{90}{$ECL$}}
    & 96  &\bf{0.166}&\bf{0.252}&0.174&\underline{0.259}&\underline{0.168}&0.272&0.197&0.282&0.198&0.276&0.193&0.308&0.201&0.317&0.274&0.368\\
    & 192 &\bf{0.172}&\bf{0.259}&\underline{0.178}&\underline{0.265}&0.184&0.289&0.196&0.285&0.198&0.279&0.201&0.315&0.222&0.334&0.296&0.386\\
    & 336 &\bf{0.191}&\bf{0.277}&\underline{0.196}&\underline{0.282}&0.198&0.300&0.209&0.301&0.217&0.301&0.214&0.329&0.254&0.361&0.300&0.394\\
    & 720 &\underline{0.231}&\bf{0.309}&0.237&\underline{0.316}&\bf{0.220}&0.320&0.245&0.333&0.279&0.357&0.246&0.355&0.254&0.361&0.373&0.439\\
    & Avg &\bf{0.189}&\bf{0.274}&0.196&\underline{0.280}&\underline{0.192}&0.295&0.212&0.300&0.223&0.303&0.214&0.327&0.227&0.338&0.311&0.397\\
    \midrule
    
    \multirow{5}{*}{\rotatebox{90}{$Traffic$}}
    & 96  &\bf{0.428}&\bf{0.276}&\underline{0.477}&\underline{0.305}&0.593&0.321&0.650&0.396&0.649&0.391&0.587&0.366&0.613&0.388&0.274&0.368\\
    & 192 &\bf{0.434}&\bf{0.274}&\underline{0.471}&\underline{0.299}&0.617&0.336&0.598&0.370&0.603&0.366&0.604&0.373&0.616&0.382&0.296&0.386\\
    & 336 &\bf{0.448}&\bf{0.282}&\underline{0.485}&\underline{0.305}&0.629&0.336&0.605&0.373&0.613&0.371&0.621&0.383&0.622&0.337&0.300&0.394\\
    & 720 &\bf{0.478}&\bf{0.299}&\underline{0.518}&\underline{0.325}&0.640&0.350&0.645&0.394&0.692&0.427&0.626&0.382&0.660&0.408&0.373&0.439\\
    & Avg &\bf{0.447}&\bf{0.282}&\underline{0.487}&\underline{0.308}&0.620&0.336&0.625&0.383&0.639&0.389&0.610&0.376&0.628&0.379&0.311&0.397\\
    \midrule
    % \multicolumn{2}{c|}{Average}&\\
    
    \bottomrule
    \end{tabular}
    }
    \end{center}
\end{table*}

\subsection{Detailed Results of Multivariate Forecasting with RBF Kernel and Polynomial Kernel}
\label{app:kernel_full}
\begin{table*}
\caption{Multivairate forecasting results with RBF kernel and polynomial kernel. The lookback length is set as 96. All models are evaluated on 4 different prediction horizons \{96, 192, 336, 720\}. A lower MSE indicates better performance. The best ones are in Bold, and the second ones are \underline{underlined}.}
\label{tab:kernel_forecasting}
\begin{center}

\scalebox{0.8}{
\begin{tabular}{c|c|cc|cc|cc|cc|cc}
\toprule
\midrule

\multicolumn{2}{c|}{Methods}&\multicolumn{2}{c|}{AttnEmbed}&\multicolumn{2}{c|}{RBF Kernel}&\multicolumn{2}{c|}{Polynomial Kernel}&\multicolumn{2}{c|}{PatchTST}&\multicolumn{2}{c|}{TimesNet} \\

\midrule

\multicolumn{2}{c|}{Metric} & MSE& MAE & MSE & MAE& MSE & MAE& MSE  & MAE& MSE  & MAE\\
\midrule

\multirow{5}{*}{\rotatebox{90}{$Weather$}}
& 96  &\bf{0.171}&\bf{0.215}&0.175&0.220&0.174&\underline{0.216}&0.178&0.219&\underline{0.172}&0.220\\
& 192 &\bf{0.218}&\bf{0.257}&0.222&0.258&0.221&\bf{0.257}&0.224&0.259&\underline{0.219}&0.261\\
& 336 &\underline{0.274}&\underline{0.297}&0.276&\underline{0.297}&\bf{0.272}&\bf{0.296}&0.278&0.298&0.280&0.306\\
& 720 &\underline{0.348}&\bf{0.346}&\bf{0.347}&\bf{0.346}&0.350&\bf{0.346}&0.350&\bf{0.346}&0.365&0.359 \\
& Avg &\bf{0.252}&\bf{0.278}&0.255&0.280&\underline{0.254}&\underline{0.279}&0.257&0.280&0.259&0.287 \\
\midrule

\multirow{5}{*}{\rotatebox{90}{$ETTh1$}}
& 96  &\bf{0.367}&\bf{0.398}&\underline{0.374}&\bf{0.398}&0.380&0.400&0.393&0.408&0.384&0.402\\
& 192 &\bf{0.420}&\bf{0.428}&0.441&0.436&0.437&0.431&0.445&0.434&\underline{0.436}&\underline{0.429}\\
& 336 &\bf{0.448}&\bf{0.438}&0.475&0.452&\underline{0.457}&\underline{0.442}&0.484&\underline{0.451}&0.491&0.469\\
& 720 &\underline{0.454}&\underline{0.459}&0.491&0.472&\bf{0.450}&\bf{0.453}&0.480&0.471&0.521&0.500\\
& Avg &\bf{0.422}&\bf{0.430}&0.445&0.439&\underline{0.431}&\underline{0.431}&0.450&0.440&0.458&0.450\\
\midrule

% \multirow{5}{*}{\rotatebox{90}{$ETTh2$}}
% & 96  &\underline{0.296}&\underline{0.346}&\bf{0.294}&\bf{0.343}&0.340&0.374&0.333&0.387&0.296&0.344&0.358&0.397&0.346&0.388&3.755&1.525\\
% & 192 &\bf{0.369}&\bf{0.392}&\underline{0.377}&\underline{0.393}&0.402&0.414&0.477&0.476&0.389&0.402&0.429&0.439&0.456&0.452&5.602&1.931\\
% & 336 &\bf{0.376}&\bf{0.405}&\underline{0.381}&\underline{0.409}&0.452&0.452&0.594&0.541&0.418&0.430&0.496&0.487&0.482&0.486&4.721&1.835\\
% & 720 &\bf{0.405}&\bf{0.432}&\underline{0.412}&\underline{0.471}&0.462&0.468&0.831&0.657&0.433&0.448&0.463&0.474&0.515&0.511&3.647&1.625\\
% & Avg &\bf{0.361}&\bf{0.393}&\underline{0.366}&\underline{0.404}&0.414&0.427&0.559&0.515&0.384&0.406&0.437&0.449&0.450&0.459&4.431&1.729\\
% \midrule

\multirow{5}{*}{\rotatebox{90}{$ETTm1$}}
& 96  &\underline{0.317}&\underline{0.356}&\bf{0.316}&\underline{0.356}&0.318&\bf{0.355}&0.321&0.360&0.338&0.375\\
& 192 &\underline{0.357}&0.381&0.358&\underline{0.380}&\bf{0.354}&\bf{0.377}&0.362&0.384&0.371&0.387\\
& 336 &\bf{0.387}&0.404&\underline{0.389}&\underline{0.403}&0.391&\underline{0.403}&0.392&\bf{0.402}&0.410&0.411\\
& 720 &\bf{0.448}&0.439&\underline{0.450}&\bf{0.435}&0.453&0.437&\underline{0.450}&\bf{0.435}&0.478&0.450\\
& Avg &\bf{0.377}&0.395&\underline{0.378}&\bf{0.393}&0.379&\bf{0.393}&0.381&0.395&0.400&0.406\\
\midrule
% \multicolumn{2}{c|}{Average}&\\

\bottomrule
\end{tabular}
}
\end{center}
\end{table*}

\subsection{Detailed Results of Utilize AttnEmbed as A Plug-in}
\label{app:plugin_full}
\begin{table}[h]
\caption{Utilize AttnEmbed as a plug-in. The lookback length is 336 for PatchTST and 96 for CARD. All models are evaluated on 4 different prediction horizons \{96, 192, 336, 720\}. A lower MSE indicates better performance.}
\label{tab:plugin}
\begin{center}
\scalebox{0.72}{
\begin{tabular}{c|c|cc|cc|cc|cc}
\toprule
\midrule

\multicolumn{2}{c|}{\multirow{2}{*}{Methods}}&\multicolumn{2}{c|}{\multirow{2}{*}{PatchTST(42)}}&\multicolumn{2}{c|}{PatchTST(42)}&\multicolumn{2}{c|}{\multirow{2}{*}{CARD}}&\multicolumn{2}{c}{CARD} \\
\multicolumn{2}{c|}{}&\multicolumn{2}{c|}{}&\multicolumn{2}{c|}{+AttnEmbed}&\multicolumn{2}{c|}{}&\multicolumn{2}{c}{+AttnEmbed} \\

\midrule

\multicolumn{2}{c|}{Metric} & MSE& MAE & MSE & MAE& MSE & MAE& MSE  & MAE\\
\midrule

\multirow{5}{*}{\rotatebox{90}{$Weather$}}
& 96  &0.152&0.199&\textbf{0.151}&\textbf{0.199}&\textbf{0.150}&\textbf{0.188}&0.152&0.190\\
& 192 &0.197&0.243&\textbf{0.195}&\textbf{0.240}&0.202&0.238&\textbf{0.200}&\textbf{0.237}\\
& 336 &0.249&0.283&\textbf{0.246}&\textbf{0.282}&0.260&0.282&\textbf{0.259}&\textbf{0.282}\\
& 720 &0.320&0.335&\textbf{0.320}&\textbf{0.331}&0.343&0.353&\textbf{0.341}&\textbf{0.334}\\
& Avg &0.229&0.265&\textbf{0.227}&\textbf{0.263}&0.239&0.265&\textbf{0.238}&\textbf{0.260}\\
\midrule

\multirow{5}{*}{\rotatebox{90}{$ETTh1$}}
& 96  &0.375&0.399&\textbf{0.374}&\textbf{0.397}&0.383&0.391&\textbf{0.379}&\textbf{0.390}\\
& 192 &0.414&0.421&\textbf{0.412}&\textbf{0.423}&0.435&0.420&\textbf{0.428}&\textbf{0.421}\\
& 336 &0.431&0.436&\textbf{0.420}&\textbf{0.432}&0.479&0.442&\textbf{0.472}&\textbf{0.440}\\
& 720 &0.449&0.466&\textbf{0.430}&\textbf{0.455}&0.471&0.461&\textbf{0.469}&\textbf{0.459}\\
& Avg &0.417&0.430&\textbf{0.409}&\textbf{0.426}&0.442&0.428&\textbf{0.436}&\textbf{0.427}\\
\midrule

\multirow{5}{*}{\rotatebox{90}{$ETTm1$}}
& 96  &0.290&0.342&\textbf{0.286}&\textbf{0.341}&0.319&0.347&\textbf{0.316}&\textbf{0.344}\\
& 192 &0.332&0.369&\textbf{0.331}&\textbf{0.369}&0.363&0.370&\textbf{0.356}&\textbf{0.365}\\
& 336 &0.366&0.392&\textbf{0.363}&\textbf{0.390}&0.392&0.390&\textbf{0.386}&\textbf{0.386}\\
& 720 &0.420&0.424&\textbf{0.410}&\textbf{0.416}&0.458&0.425&\textbf{0.450}&\textbf{0.422}\\
& Avg &0.352&0.381&\textbf{0.347}&\textbf{0.379}&0.383&0.384&\textbf{0.377}&\textbf{0.379}\\
\midrule

\bottomrule
\end{tabular}
}

\end{center}
\end{table}

\subsection{Detailed Results of Utilize AttnEmbed as A Plug-in}
\label{app:ablation_full}
\begin{table}[h]
\caption{Ablation on EMA and landmark. The lookback length is 96. All models are evaluated on 4 different prediction horizons \{96, 192, 336, 720\}. A lower MSE indicates better performance.}
\label{tab:ablation_ema_landmark}
\begin{center}
\scalebox{0.8}{

\begin{tabular}{c|c|cc|cc|cc}
\toprule
\midrule

\multicolumn{2}{c|}{\multirow{2}{*}{Methods}}&\multicolumn{2}{c|}{\multirow{2}{*}{AttnEmbed}}&\multicolumn{2}{c|}{AttnEmbed}&\multicolumn{2}{c}{AttnEmbed} \\
\multicolumn{2}{c|}{}&\multicolumn{2}{c|}{}&\multicolumn{2}{c|}{w/o EMA }&\multicolumn{2}{c}{w/o Landmark} \\

\midrule

\multicolumn{2}{c|}{Metric} & MSE& MAE & MSE & MAE& MSE & MAE\\
\midrule

\multirow{5}{*}{\rotatebox{90}{$ETTh1$}}
& 96  &0.367&0.398&0.378&0.398&0.370&0.400\\
& 192 &0.420&0.428&0.423&0.425&0.421&0.427\\
& 336 &0.448&0.438&0.456&0.430&0.440&0.436\\
& 720 &0.454&0.459&0.472&0.455&0.461&0.454\\
& Avg &0.422&0.430&0.432&0.427&0.423&0.429\\
\midrule

\multirow{5}{*}{\rotatebox{90}{$ETTm1$}}
& 96  &0.317&0.356&0.319&0.360&0.326&0.365\\
& 192 &0.357&0.381&0.374&0.391&0.370&0.384\\
& 336 &0.387&0.404&0.396&0.402&0.396&0.403\\
& 720 &0.448&0.439&0.449&0.437&0.455&0.436\\
& Avg &0.377&0.395&0.384&0.397&0.386&0.397\\
\midrule

\bottomrule
\end{tabular}
}
\end{center}
\end{table}
%%\onecolumn
%%\section{You \emph{can} have an appendix here.}

%%You can have as much text here as you want. The main body must be at most $8$ pages long.
%%For the final version, one more page can be added.
%%If you want, you can use an appendix like this one.  

%%The $\mathtt{\backslash onecolumn}$ command above can be kept in place if you prefer a one-column appendix, or can be removed if you prefer a two-column appendix.  Apart from this possible change, the style (font size, spacing, margins, page numbering, etc.) should be kept the same as the main body.
%%%%%%%%%%%%%%%%%%%%%%%%%%%%%%%%%%%%%%%%%%%%%%%%%%%%%%%%%%%%%%%%%%%%%%%%%%%%%%%
%%%%%%%%%%%%%%%%%%%%%%%%%%%%%%%%%%%%%%%%%%%%%%%%%%%%%%%%%%%%%%%%%%%%%%%%%%%%%%%

\end{document}